\documentclass[journal]{IEEEtranTIE}
\usepackage{graphicx}
\graphicspath{{figures/}}
\usepackage{grffile} 
\usepackage{afterpage} 
\usepackage{cite}
\usepackage{amsmath}
\usepackage{amssymb}
\usepackage{amsthm}
\usepackage{mathtools}
\usepackage{bm}
\usepackage{url}

\usepackage[latin1]{inputenc}
\usepackage{colortbl}
\usepackage{soul}
\usepackage{multirow}
\usepackage{pifont}
\usepackage{color}
\usepackage{alltt}
\usepackage[hidelinks]{hyperref}
\hypersetup{hypertexnames=false}
\usepackage{enumerate}
\usepackage{siunitx}
\usepackage{breakurl}
\usepackage{epstopdf}
\usepackage{pbox}
\usepackage{algorithm}
\usepackage{algorithmic}
\usepackage{float}
\usepackage{booktabs}
\usepackage{placeins}

\theoremstyle{definition}

\theoremstyle{remark}
\newtheorem{remark}{Remark}

\begin{document}

\title{Time-Efficient Iterative Learning Planning for Safety-Critical Dynamic Obstacle Avoidance}

\author{\sffamily Zhiyi Chen, Shuli Lv, Chen Min, Yong Xu, Jian Sun,~\textit{Senior Member,~IEEE},\\
and Quan Quan,~\textit{Senior Member,~IEEE}%
\thanks{Corresponding author: Yong Xu.}%
\thanks{Zhiyi Chen, Yong Xu, and Jian Sun are with the National Key Laboratory of Autonomous Intelligent Unmanned Systems, School of Automation, Beijing Institute of Technology, Beijing 100081, China (e-mail: chenzhiyi@bit.edu.cn; xuyong@bit.edu.cn; sunjian@bit.edu.cn).}%
\thanks{Shuli Lv, Chen Min, and Quan Quan are with the School of Automation Science and Electrical Engineering, Beihang University, Beijing 100191, China (e-mail: lvshuli@buaa.edu.cn; min\_chen@buaa.edu.cn; qq\_buaa@buaa.edu.cn).}%
}


\maketitle
\newcommand{\ResetBodyFootnoteSpacing}{\global\skip\footins=0.9\baselineskip plus 0.4\baselineskip minus 0.2\baselineskip\relax\global\footnotesep=0.8\baselineskip}
\ifdefined\AddToHookNext
  \AddToHookNext{shipout/after}{\ResetBodyFootnoteSpacing}
\else
  \afterpage{\ResetBodyFootnoteSpacing}
\fi
\setlength{\parskip}{0pt}

\begin{abstract}
\boldmath
Autonomous mobile robots require time-efficient planning and safety-critical dynamic obstacle avoidance under constrained onboard computation. While Iterative Learning Planning (ILP) offers lightweight and efficient traversal planning, it lacks explicit mechanisms for dynamic obstacle perception and avoidance. This article extends ILP to safety-critical navigation in dynamic environments by integrating an anticipatory risk-blended control barrier function (ARB-CBF). The extended ILP learns traversal-speed and steering-bias profiles via a fractional-power update based on local obstacle risk, generating nominal control commands that ARB-CBF modifies at runtime for real-time safety guarantees. Algorithmic analysis demonstrates that the ILP replanning stage scales at $O(kN)$ for $k$ iterations and $N$ waypoints, while ARB-CBF executes with linear complexity. Comprehensive simulations and real-world experiments validate the framework, demonstrating superior temporal efficiency and safety with lower computational overhead compared to optimization-based baselines, making it highly suitable for resource-constrained platforms.
\end{abstract}



\section{Introduction}
\label{sec:introduction}

\begin{figure}[!t]
\centering
\includegraphics[width=\columnwidth]{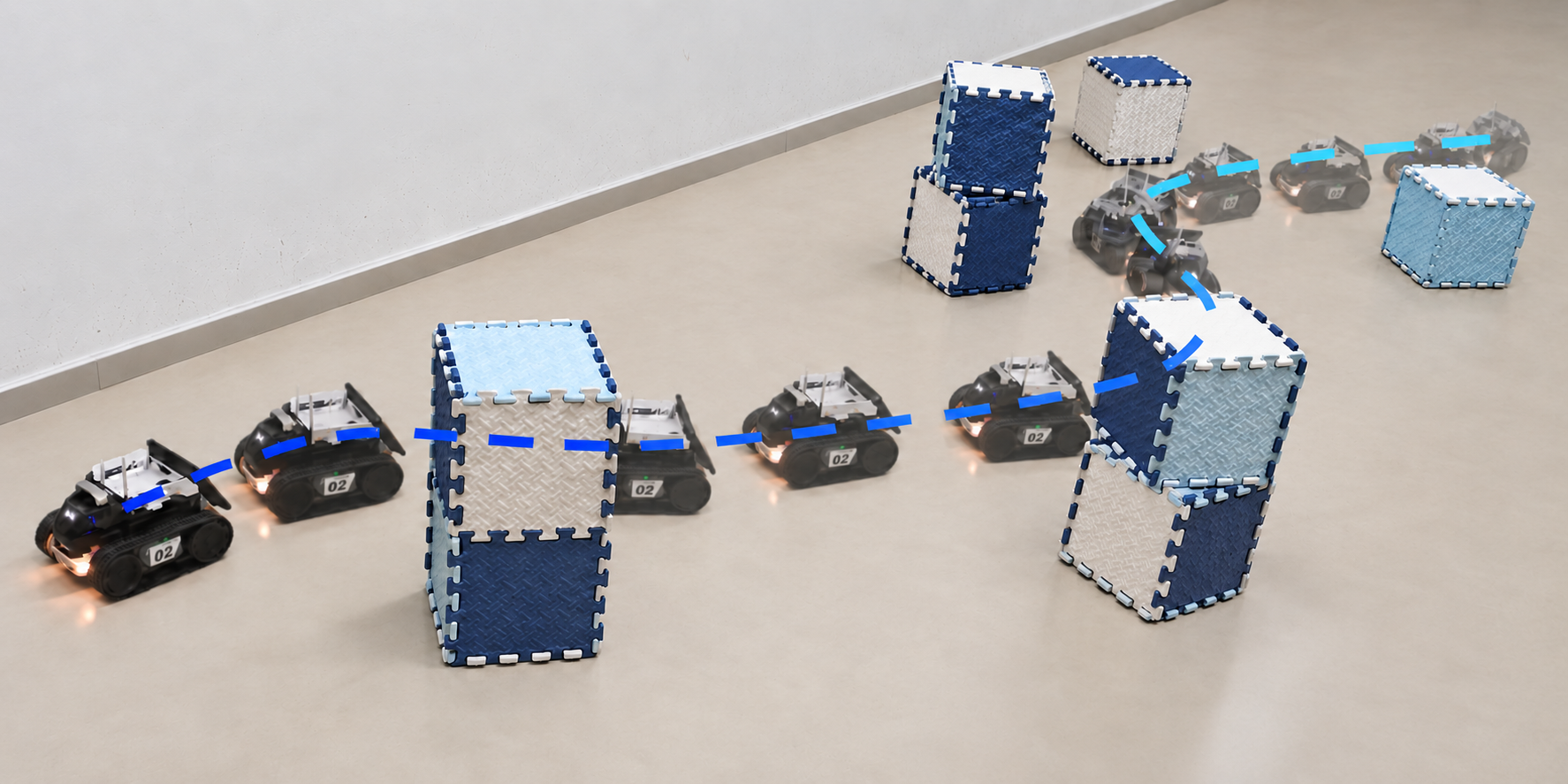}

\caption{Application of the proposed methodology to a wheeled mobile robot for planning safe and time-efficient trajectories in an unknown environment.}
\label{fig:application}
\end{figure}

Autonomous mobile robots (AMRs) are being rapidly deployed in industrial transportation, automated inspection, and service applications. As illustrated in Fig.~\ref{fig:application}, navigating safely and efficiently through dynamic environments remains a prerequisite for these autonomous tasks~\cite{guiochet2017safety}. Effective motion planning requires a delicate trade-off among traversal efficiency, collision avoidance, and online computational overhead~\cite{gonzalez2016review}. While aggressive control strategies minimize traversal time, they severely compromise reaction margins under sudden dynamic obstacles. Conversely, overly conservative avoidance and frequent global replanning impair transport efficiency and strain onboard processing capabilities. Consequently, enabling time-efficient traversal and safety-critical avoidance under tight computational constraints remains a critical challenge for resource-limited mobile platforms.

To address these limitations, local planning frameworks aim to synthesize collision-free trajectories using real-time sensory feedback. Reactive methods, including the Dynamic Window Approach (DWA)~\cite{su2025dwa}, Artificial Potential Fields (APFs), Reciprocal Velocity Obstacles (RVO)~\cite{vanDenBerg2008rvo}, and Generalized Velocity Obstacles (GVO)~\cite{li2025gvo,refYang}, offer computationally lightweight obstacle avoidance. However, these purely reactive strategies often exhibit conservative behaviors, susceptibility to local minima, or erratic motion profiles when operating in highly dynamic, unpredictable environments.

Beyond reactive schemes, optimization-based motion planners synthesize trajectories by explicitly modeling dynamic constraints and environment bounds~\cite{paden2016survey}. For safety-critical navigation, Control Barrier Functions (CBFs) offer a mathematically rigorous mechanism to guarantee the forward invariance of safe sets~\cite{jian2023dynamic}. Integrating CBF constraints into Model Predictive Control (MPC) has demonstrated compelling performance in balancing state boundaries, actuator limits, and dynamic obstacle avoidance~\cite{zeng2021safety}. Recent advances, including dynamic CBF formulations, dual-filter safety schemes, and multi-constraint CBF variants, have further enhanced safety-critical navigation in dynamic environments~\cite{jian2023dynamic,zhang2026extended}. However, these optimization-based CBF frameworks typically mandate solving non-convex or Quadratic Programs (QPs) at every control cycle, scaling with \(O(n^2)\) complexity (or higher) relative to the optimization dimension \(n\). This computational burden rapidly escalates with longer prediction horizons and dense obstacle distributions. Consequently, there is a strong demand for lightweight, safety-critical navigation frameworks capable of real-time obstacle avoidance without relying on computationally expensive online optimization.

Leveraging historical execution data, iterative learning control has emerged as an effective paradigm to achieve low-computational control by updating feedforward inputs across iterations~\cite{saab2022iterative,zhou2023tracking}. Predictive iterative controllers have also been formulated for real-time trajectory tracking under vehicle kinematic and dynamic constraints~\cite{wang2025constrained}. Nevertheless, these conventional methods focus exclusively on tracking pre-generated references, meaning overall traversal efficiency remains strictly bottlenecked by the initial planner. To bridge this gap, Iterative Learning Planning (ILP) shifts the iterative refinement paradigm to the planning level~\cite{lv2025high,lv2026timeoptimal}, continuously optimizing motion profiles across trials without incurring the computational overhead of large-scale online optimization. Recent literature has integrated safety-aware learning MPC and low-complexity vector-field ILP schemes to accelerate motion generation~\cite{wang2026offline,zhang2025asv}. However, existing ILP strategies predominantly assume static bounds or pre-defined corridors, lacking explicit mechanisms to guarantee real-time safety under unpredictable dynamic obstacle interactions.

Building upon our previous research on spatial iterative learning within virtual tubular boundaries for time-optimal and energy-optimal navigation~\cite{lv2025high,lv2026timeoptimal,min2026energy}, this article addresses the unhandled dynamic safety risks. While our prior studies demonstrated that model-free ILP significantly improves traversal efficiency with low computational overhead, they primarily considered nominal trajectories within static collision-free corridors, leaving the system vulnerable to dynamic obstacles entering the virtual tube.

To address these limitations, the proposed framework extends the ILP algorithm in both iterative learning and runtime safety. In contrast to prior work~\cite{lv2025high}, the main extensions are threefold: (1) single-channel learning is extended to jointly learn traversal speed and steering bias, with a fractional-power update improving finite-iteration performance; (2) local dynamic obstacle perception and risk evaluation are incorporated for risk-aware replanning; (3) this article is the first to introduce a control barrier function into the ILP framework for runtime speed and steering correction, enabling dynamic obstacle avoidance in real time. These extensions retain lightweight, time-efficient planning while enabling safety-critical navigation in dynamic environments.

The principal contributions of this article are threefold:
\vspace{-0.5em}
\begin{enumerate}
\item \textbf{Computationally Lightweight and Time-Efficient Planning:}
We propose a Multi-Input Multi-Output Fractional-Power Update Rule ILP (MIMO-FPUR-ILP) framework that jointly adapts traversal-speed and steering-bias profiles under local risk feedback. The fractional-power update mechanism significantly accelerates passing-time reduction within few iterations, maintaining a model-free, low-complexity architecture.
\item \textbf{Learning-Compatible Safety-Critical Correction:}
We design an anticipatory safety filter that seamlessly combines with ILP. The filter preserves learned reference profiles during unconstrained operation and intervenes under impending collision risks. Leveraging LiDAR and IMU measurements, an explicit CBF formulation ensures real-time dynamic obstacle avoidance without global obstacle information or repeated online optimization.
\item \textbf{Comprehensive Experimental Validation:}
Extensive simulations and real-world experiments demonstrate the superiority of the proposed framework. The algorithm achieves an over \(90\%\) reduction in online planning overhead compared to state-of-the-art optimization-based baselines, while simultaneously enhancing traversal efficiency and safety in dynamic scenarios.
\end{enumerate}

\section{Preliminaries and Problem Formulation}
\label{sec:preliminaries}

\subsection{Virtual Tube and Path Parameterization}

In previous work, the virtual tube was used for robotic path planning~\cite{mao2025optimal}. The key insight is to replace explicit avoidance of individual static obstacles with containment inside a precomputed safe corridor.

\begin{figure}[!t]\centering

\includegraphics[width=\columnwidth]{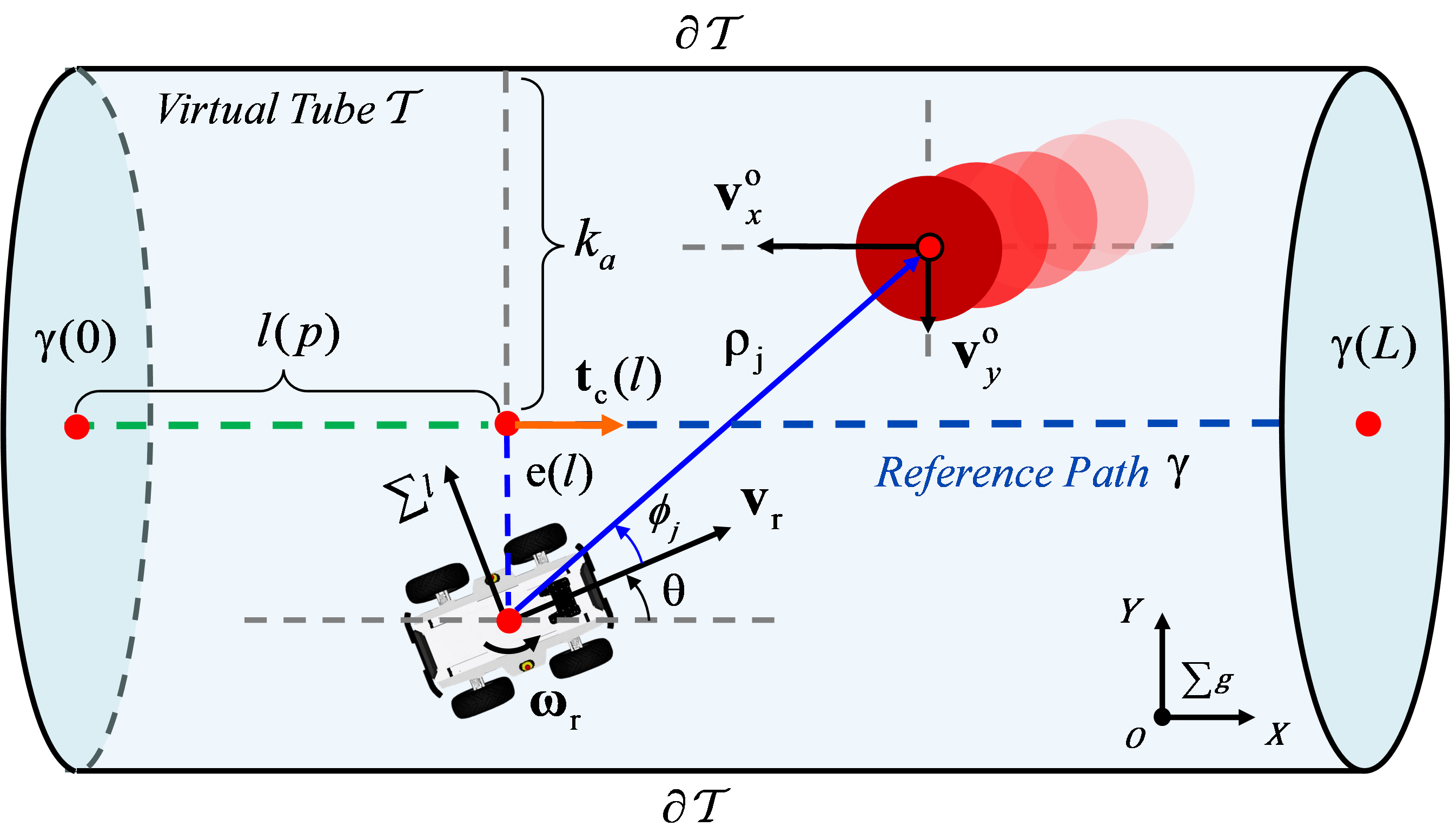}

\caption{Local polar frame and arc length parameterization within the virtual tube.}\label{fig:virtual_tube}

\end{figure}

As illustrated in Fig.~\ref{fig:virtual_tube}, a collision-free path connecting start and goal is chosen as the generating curve $\gamma: [0, L] \to \mathbb{R}^2$. The virtual tube $\mathcal{T}$ of radius $k_a > 0$ is then defined by
\begin{equation}
\mathcal{T} = \left\{ \mathbf{p} \in \mathbb{R}^2 \;\big|\; \big\| \mathbf{p} - \gamma(l) \big\|_\perp \leq k_a, \; l \in [0, L] \right\},
\label{eq:virtual_tube}
\end{equation}
where $l\in[0,L]$ is the arc length coordinate and $\|\cdot\|_\perp$ denotes perpendicular distance. The radius $k_a$ is selected such that $\mathcal{T}$ lies within the static free space; thus, trajectories remaining inside the tube avoid the static obstacles considered during global planning.

To establish a path indexed spatial representation, the generator curve is sampled at $N$ arc length waypoints. For a robot position $\mathbf p$, it is projected onto the closest arc length coordinate $l=l(\mathbf p)=\arg\min_{s\in[0,L]}\|\mathbf p-\gamma(s)\|$, as illustrated in Fig.~\ref{fig:virtual_tube}. Let $\mathbf t_c(l)=\dot{\gamma}(l)$ denote the unit tangent. The signed cross track error is defined as
\begin{equation}
e(l)=\bigl(\mathbf p-\gamma(l)\bigr)^\top\mathbf R(\pi/2)\mathbf t_c(l),
\label{eq:cross_track}
\end{equation}
where $\mathbf R(\pi/2)$ denotes the $90^\circ$ rotation matrix. For $v_r(l)>0$, $\mathrm{d}t=\mathrm{d}l/v_r(l)$ expresses traversal time in the spatial domain.

\subsection{System Model}

\subsubsection{Vehicle Model}

The robot is modeled as a differential drive mobile platform. Let $\mathbf{x} = [\mathbf{p}^\top, \theta]^\top$ denote the system state with position $\mathbf{p} = [x, y]^\top$ and orientation $\theta$, and let $\mathbf{u}_r = [v_r, \omega_r]^\top$ denote the robot's linear and angular velocities, measurable via onboard IMU. The kinematic model in the local frame $\Sigma_l$ is written as
\begin{equation}
\dot{\mathbf{x}} = \begin{bmatrix} \dot{x} \\ \dot{y} \\ \dot{\theta} \end{bmatrix} =
\begin{bmatrix} \cos\theta & 0 \\ \sin\theta & 0 \\ 0 & 1 \end{bmatrix}
\begin{bmatrix} v_r \\ \omega_r \end{bmatrix},
\label{eq:diff_drive}
\end{equation}
with $0 \leq v_r \leq v_{\max}$ and $|\omega_r| \leq \omega_{\max}$.

\subsubsection{Obstacle Model}

For obstacle avoidance, it is convenient to describe the relative geometry between the robot and each obstacle in a local polar frame, as shown in Fig.~\ref{fig:virtual_tube}. Let $\mathbf{p}_j^o=[x_j^o,y_j^o]^\top$ and $\mathbf{v}_j^o=[v_{x,j}^o,v_{y,j}^o]^\top$ denote the position and velocity of the $j$-th obstacle in $\Sigma_g$. The relative distance $\rho_j$ and bearing angle $\phi_j$ in the robot's local frame are defined as
\begin{align}
\rho_j &= \sqrt{(x_j^{\rm o} - x)^2 + (y_j^{\rm o} - y)^2}, \\
\phi_j &= \operatorname{atan2}(y_j^{\rm o} - y, x_j^{\rm o} - x) - \theta.
\label{eq:relative_state}
\end{align}
Here, $\rho_j$ denotes the obstacle distance, and $\phi_j$ denotes the obstacle bearing measured from the robot heading. The static safety radius is then defined as
\begin{equation}
R_{0,j} = r_r + r_j^{\rm o} + d_0,
\label{eq:R0}
\end{equation}
where $r_r$ and $r_j^{\rm o}$ are the circumscribed radii of the robot and the $j$-th obstacle, and $d_0 > 0$ is a static clearance margin. The obstacle motion is modeled as a single integrator $\dot{\mathbf{p}}_j^{\rm o} = \mathbf{v}_j^{\rm o}$ with bounded velocity $\|\mathbf{v}_j^{\rm o}\| \leq \bar{v}^{\rm o}$. Differentiating~(\ref{eq:relative_state}) along the system trajectories, the relative distance dynamics are obtained as
\begin{equation}
\dot{\rho}_j = -v_r \cos\phi_j + \mathbf{b}_j^\top \mathbf{v}_j^{\rm o},
\label{eq:rho_dot}
\end{equation}
where $\mathbf{b}_j = [\cos(\theta+\phi_j), \sin(\theta+\phi_j)]^\top$ is the unit line-of-sight vector. The term $-v_r\cos\phi_j$ captures the effect of the robot's own motion, while $\mathbf{b}_j^\top \mathbf{v}_j^{\rm o}$ is the obstacle velocity projected onto the line of sight.

\subsection{Dynamic Perception and Risk Evaluation}

\label{sec:dynamic_perception}

The onboard LiDAR point cloud is clustered via DBSCAN into circular obstacles with center $\mathbf p_m^{\rm o}$ and radius $r_m^{\rm o}$. Nearest-neighbor association and a Kalman filter are used to maintain temporal consistency and estimate the obstacle velocity $\mathbf{v}_m^{\rm o}$.

\begin{figure}[!t]
\centering
\begin{minipage}[c]{0.35\columnwidth}
\centering
\includegraphics[width=\linewidth]{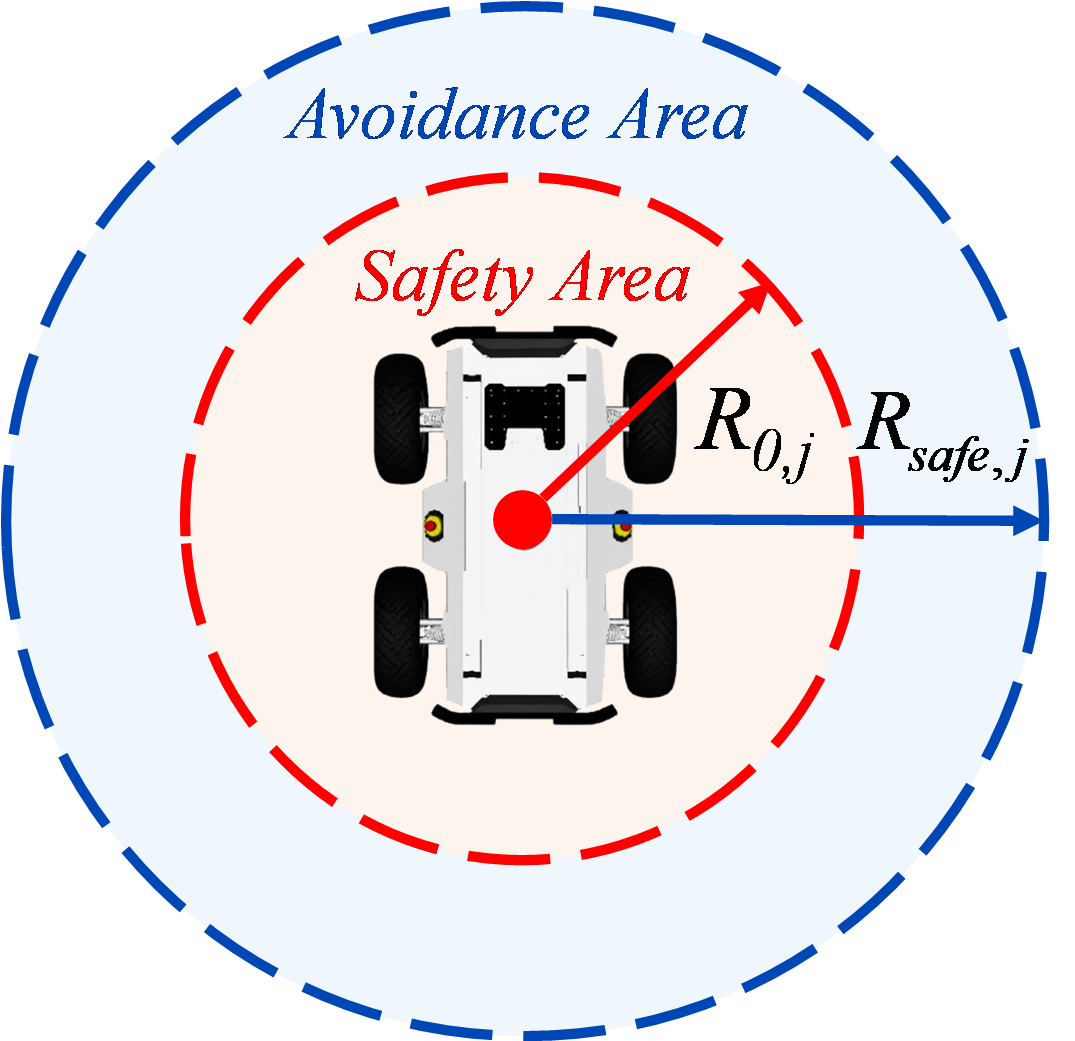}
\end{minipage}\hfill
\begin{minipage}[c]{0.61\columnwidth}
\centering
\includegraphics[width=\linewidth]{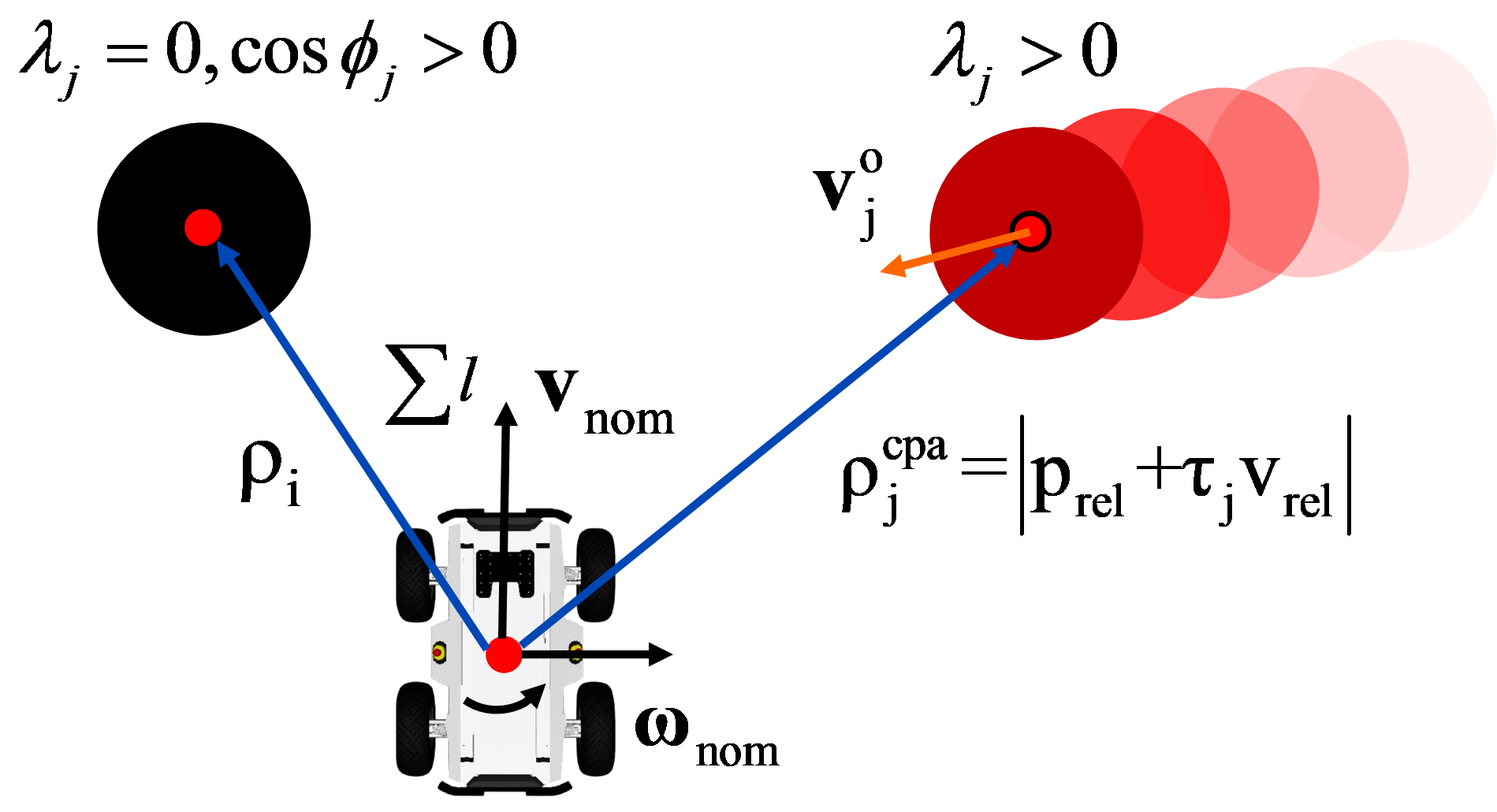}
\end{minipage}\\[1.2mm]
\makebox[0.35\columnwidth][c]{\footnotesize(a)}\hfill
\makebox[0.61\columnwidth][c]{\footnotesize(b)}
\caption{Dynamic perception and risk evaluation. (a) Safety and response boundaries. (b) Relative-motion risk evaluation with CPA prediction.}
\label{fig:dynamic_risk}
\end{figure}

Obstacles with $\|\mathbf{v}_m^{\rm o}\|\leq v_0$ are classified as static. For moving obstacles, a short-horizon relative-motion prediction is performed. Let $\mathbf p_{\rm rel}=\mathbf p-\mathbf p_j^{\rm o}$ and $\mathbf{v}_{\rm rel}=v_{\rm nom}[\cos\theta,\sin\theta]^\top-\mathbf{v}_j^{\rm o}$, where $v_{\rm nom}$ is the nominal forward speed. The predicted closest-approach time is
\begin{equation}
\tau_j=\Pi_{[0,\tau_{\max}]}\!\left(-\frac{\mathbf p_{\rm rel}^\top\mathbf{v}_{\rm rel}}{\|\mathbf{v}_{\rm rel}\|^2+\varepsilon}\right),
\label{eq:cpa_time}
\end{equation}
where $\Pi_{[0,\tau_{\max}]}(\cdot)$ bounds the prediction horizon and $\varepsilon>0$ avoids numerical singularity. The corresponding closest-point-of-approach (CPA) distance is
\begin{equation}
\rho_j^{\rm cpa}=\|\mathbf p_{\rm rel}+\tau_j\mathbf{v}_{\rm rel}\|,
\label{eq:cpa_distance}
\end{equation}
which represents the predicted minimum clearance under constant-velocity extrapolation. Fig.~\ref{fig:dynamic_risk}(b) illustrates the relative-motion geometry and CPA prediction used for moving obstacles.

To account for robot and obstacle motion, the velocity-adaptive response radius is defined as
\begin{equation}
R_{{\rm safe},j}=R_{0,j}+\kappa|v_{\rm nom}|+\eta\|\mathbf{v}_j^{\rm o}\|,
\label{eq:safe_radius}
\end{equation}
where $\kappa,\eta\geq0$ are look-ahead gains. Unlike the hard safety radius $R_{0,j}$, $R_{{\rm safe},j}$ defines the anticipatory response boundary, as illustrated in Fig.~\ref{fig:dynamic_risk}(a).

To evaluate obstacle risk consistently for both static and moving obstacles, the measured clearance $\rho_j$ is used for static obstacles satisfying $\|\mathbf{v}_j^{\rm o}\|\leq v_0$, while the CPA distance $\rho_j^{\rm cpa}$ is used otherwise; the resulting risk-evaluation distance is denoted by $\rho_j^{\rm r}$, where $v_0$ separates static and dynamic obstacles. The normalized risk margin, obstacle-wise risk weight, and aggregate risk are then defined as
\begin{subequations}\label{eq:risk_variables}
\begin{gather}
s_j=\frac{\rho_j^{\rm r}-R_{{\rm safe},j}}{r_0},
\label{eq:risk_margin}\\
\lambda_j=
\begin{cases}
1,&s_j\leq0,\\
1-3s_j^2+2s_j^3,&0<s_j<1,\\
0,&s_j\geq1,
\end{cases}
\label{eq:lambda}\\
W_{\rm risk}=
\begin{cases}
\max_{j\in\mathcal J_{\rm r}}\lambda_j,&\mathcal J_{\rm r}\neq\varnothing,\\
0,&\mathcal J_{\rm r}=\varnothing.
\end{cases}
\label{eq:overall_risk}
\end{gather}
\end{subequations}
Here, $r_0$ is the blending width and $\mathcal J_{\rm r}=\{j\mid\lambda_j>0\}$ is the risk obstacle set. Thus, $W_{\rm risk}\in[0,1]$ denotes the strongest perceived interaction risk. During the $k$th learning or replanning rollout, $W_{\rm risk}(t)$ is mapped to the arc-length coordinate as $W_k(l)\in[0,1]$, providing a shared risk representation for iterative learning and runtime safety correction.
\subsection{Problem Formulation}

Given a reference path $\mathcal{P}$ enclosed by the virtual tube $\mathcal{T}$, the objective is to minimize the traversal time while satisfying the tube, control, and dynamic-obstacle safety constraints. Let $\mathbf{u}_{\rm p}(l)=[v_h(l),\omega_h(l)]^\top\in\mathcal{U}_{\rm p}$ denote the spatial motion profile and $\mathbf{u}_{\rm r}(t)\in\mathcal{U}_{\rm r}$ the executed control command. The overall navigation problem is formulated as
\begin{subequations}
\label{eq:overall_problem}
\begin{align}
\mathbf{P}:\quad
\min_{\mathbf{u}_{\rm p}(\cdot),\,\mathbf{u}_{\rm r}(\cdot)}
T
&=
\int_{0}^{t_f}\mathrm{d}t
=
\int_{0}^{L}\frac{1}{v_h(l)}\,\mathrm{d}l,
\label{eq:overall_objective}
\\
&\mathbf{u}_{\rm p}(l)\in\mathcal{U}_{\rm p},
\qquad
\mathbf{u}_{\rm r}(t)\in\mathcal{U}_{\rm r},
\label{eq:input_constraints}
\\
&\operatorname{dist}\!\left(\mathbf{p}(t),\mathcal{O}_j(t)\right)\geq R_{0,j}.
\label{eq:obstacle_constraint}
\end{align}
\end{subequations}

Equation~\eqref{eq:obstacle_constraint} holds for all $t\in[0,t_f]$ and $j\in\mathcal{J}(t)$, where $\mathcal{J}(t)$ denotes the set of safety-relevant perceived obstacles. It requires the robot to maintain a clearance of at least $R_{0,j}$ from each perceived obstacle throughout the traversal. Problem~$\mathbf{P}$ therefore represents time-optimal navigation subject to control and dynamic obstacle safety constraints.

\section{Motion Planning Algorithm Design}
\label{sec:method}

\subsection{Problem Decomposition and Framework Overview}

\begin{figure}[!t]\centering
\includegraphics[width=\columnwidth]{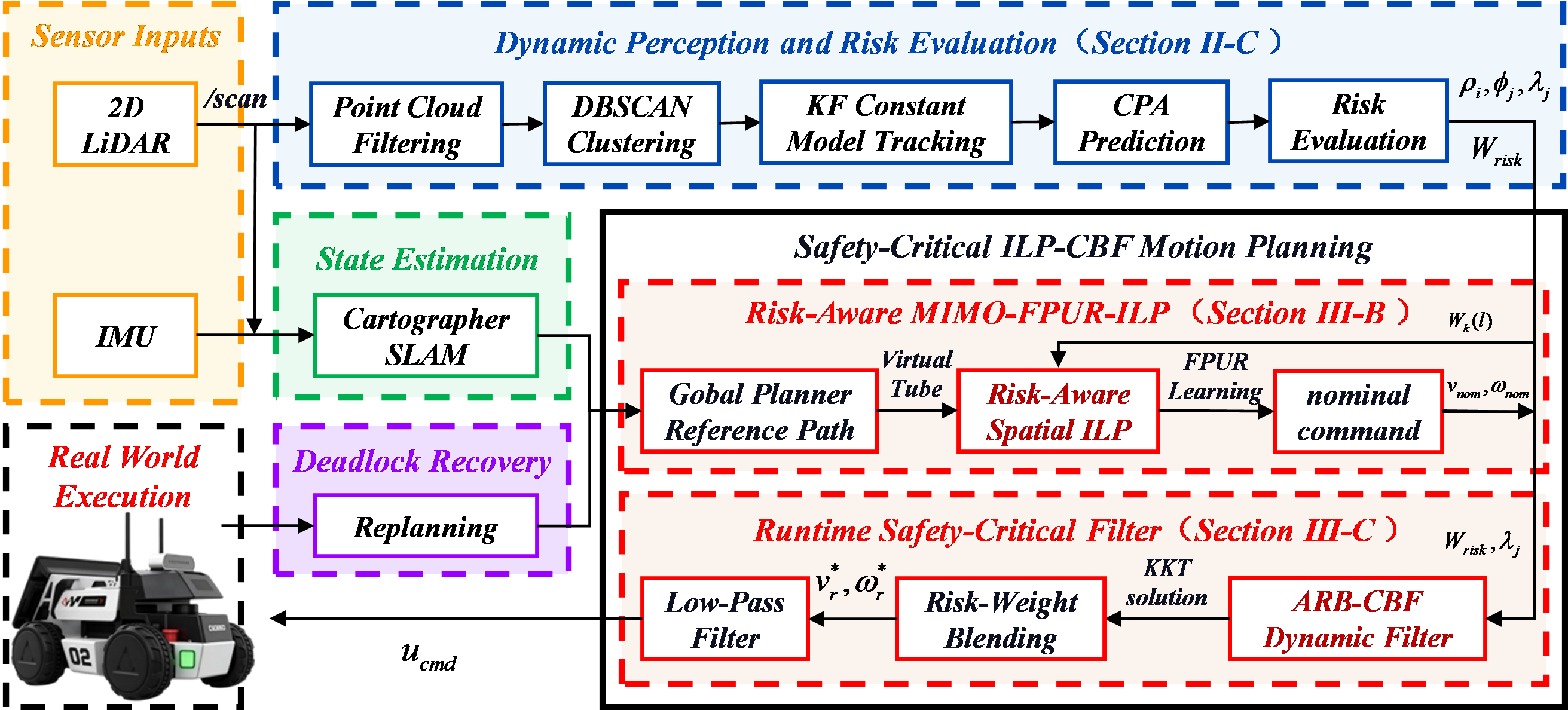}
\vspace{-0.5\baselineskip}
\caption{Overall framework of the proposed safety-critical system.}
\label{fig:framework}
\end{figure}

As shown in Fig.~\ref{fig:framework}, a global planner provides the reference path and virtual tube. Local perception estimates obstacle states and predicts short-horizon interactions with dynamic obstacles. Directly solving Problem $\mathbf{P}$ requires jointly optimizing the spatial motion profile and runtime control under time-varying safety constraints, which may introduce considerable online computation. Therefore, $\mathbf{P}$ is decomposed into two coordinated subproblems.

\textit{1) $\mathbf{P}_1$: Time-Optimal Motion Planning:} MIMO-FPUR-ILP (Section~III-B) learns bounded traversal-speed and steering-bias profiles within the virtual tube to minimize traversal time. When significant risk is perceived, the ILP update is further modulated through risk-aware replanning.

\textit{2) $\mathbf{P}_2$: Safety-Critical Dynamic Obstacle Avoidance:} Given the nominal command generated by $\mathbf{P}_1$, ARB-CBF (Section~III-C) uses local obstacle prediction and risk information to provide runtime speed and steering correction while satisfying the dynamic-obstacle safety constraints.

The two subproblems are coordinated through the nominal command and local obstacle perception. This decomposition avoids repeatedly solving the coupled Problem $\mathbf{P}$ online while retaining time-efficient motion and lightweight safety-critical dynamic obstacle avoidance.
\subsection{Time Optimal Spatial Iterative Learning Planning with Fractional Power Update Rule (FPUR)}
\label{sec:spatial_ilc}

\subsubsection{Human Racing Inspired Spatial Vector Field}

\begin{figure}[!t]\centering
\includegraphics[width=\columnwidth]{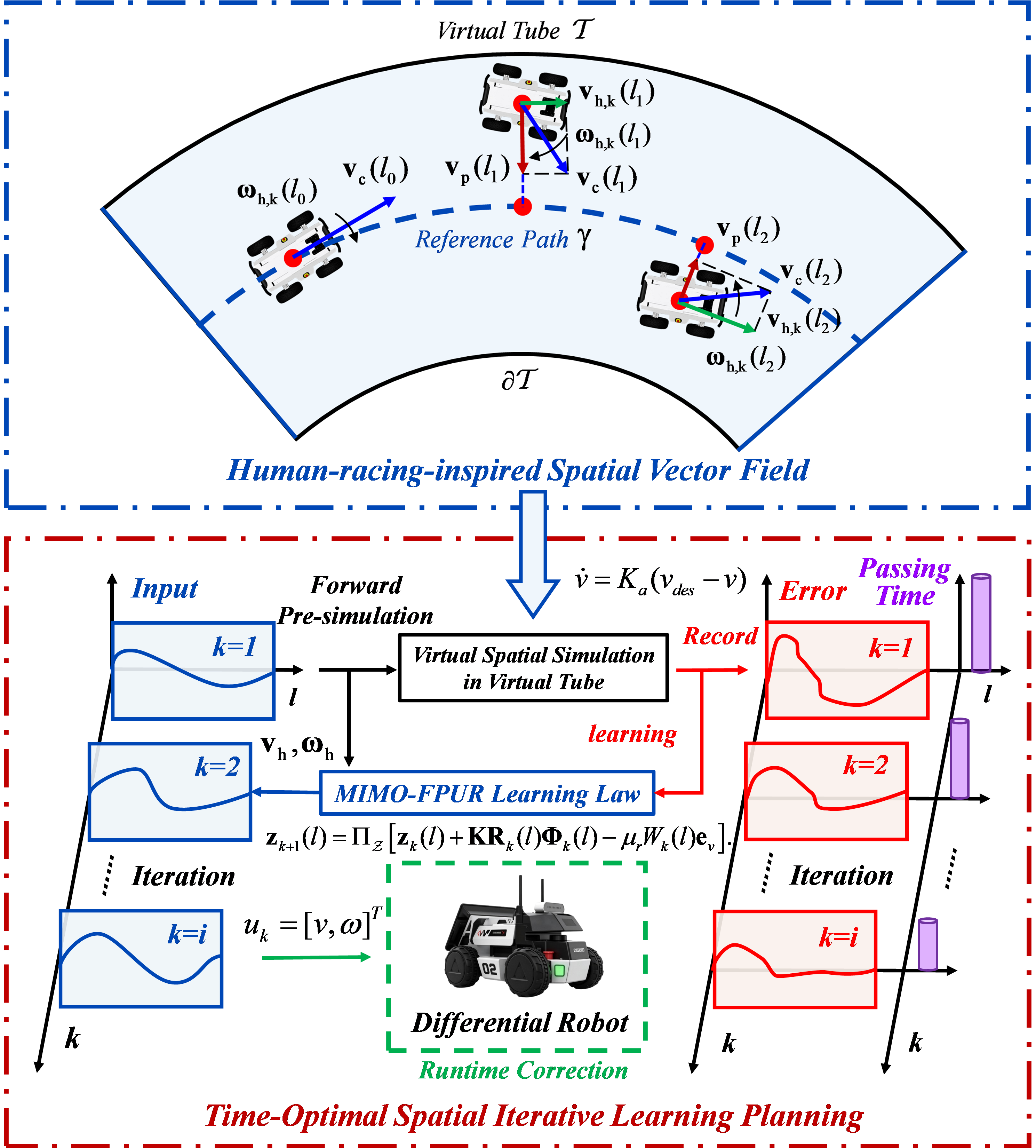}
\vspace{-0.5\baselineskip}
\caption{Workflow of time optimal spatial iterative learning planning.}\label{fig:racing_vf}
\end{figure}

Following our previous work~\cite{lv2026timeoptimal,lv2025high}, we formulate a path-indexed spatial vector field in the arc-length domain for iterative learning, as shown in Fig.~\ref{fig:racing_vf}. The spatial velocity $\mathbf{v}_c(l)$ is decomposed into convergence and traversal components as
\begin{equation}
\mathbf{v}_c(l)=\mathbf{v}_p(l)+\mathbf{v}_h(l).
\label{eq:spatial_velocity}
\end{equation}
The safety constraint $|e(l)|<k_a$ is enforced by a barrier Lyapunov function (BLF). For the signed cross-track error $e(l)$, the BLF and its spatial gradient are defined as
\begin{subequations}\label{eq:blf}
\begin{gather}
V_b(e)=\frac{1}{2}\ln\frac{k_a^2}{k_a^2-e^2},\qquad |e|<k_a,\label{eq:blf_value}\\
\nabla_{\mathbf p}V_b(e)=\frac{e}{k_a^2-e^2}\mathbf n_c(l),\label{eq:blf_gradient}
\end{gather}
\end{subequations}
where $\mathbf n_c(l)$ is the unit path normal. As established in our previous work~\cite{lv2026timeoptimal,lv2025high}, $V_b$ and its gradient grow unbounded as $|e|\rightarrow k_a$, enforcing tube-constrained tracking.

The convergence component is designed as
\begin{equation}
\mathbf{v}_p(l)=-k_1(l)\nabla_{\mathbf p}V_b(e(l)),
\label{eq:convergence_component}
\end{equation}
where $k_1(l)=k_2+k_3|\kappa(l)|$, with $k_2>0$ and $k_3\geq0$ denoting the nominal and curvature-dependent convergence gains, respectively. The traversal component is defined as
\begin{equation}
\mathbf{v}_h(l)=v_h(l)\mathbf t_c(l),
\label{eq:traversal_component}
\end{equation}
where $\mathbf t_c(l)$ is the unit path tangent and $v_h(l)\geq0$ is the adjustable traversal speed. This allocation mimics a human racing strategy, prioritizing convergence in high-curvature or poorly tracked segments while allowing faster traversal in well-tracked low-curvature segments~\cite{lv2025high}.

\begin{figure*}[!t]
\centering
\begin{minipage}[t]{0.48\textwidth}
\vspace{0pt}
\centering
\includegraphics[width=0.8\linewidth]{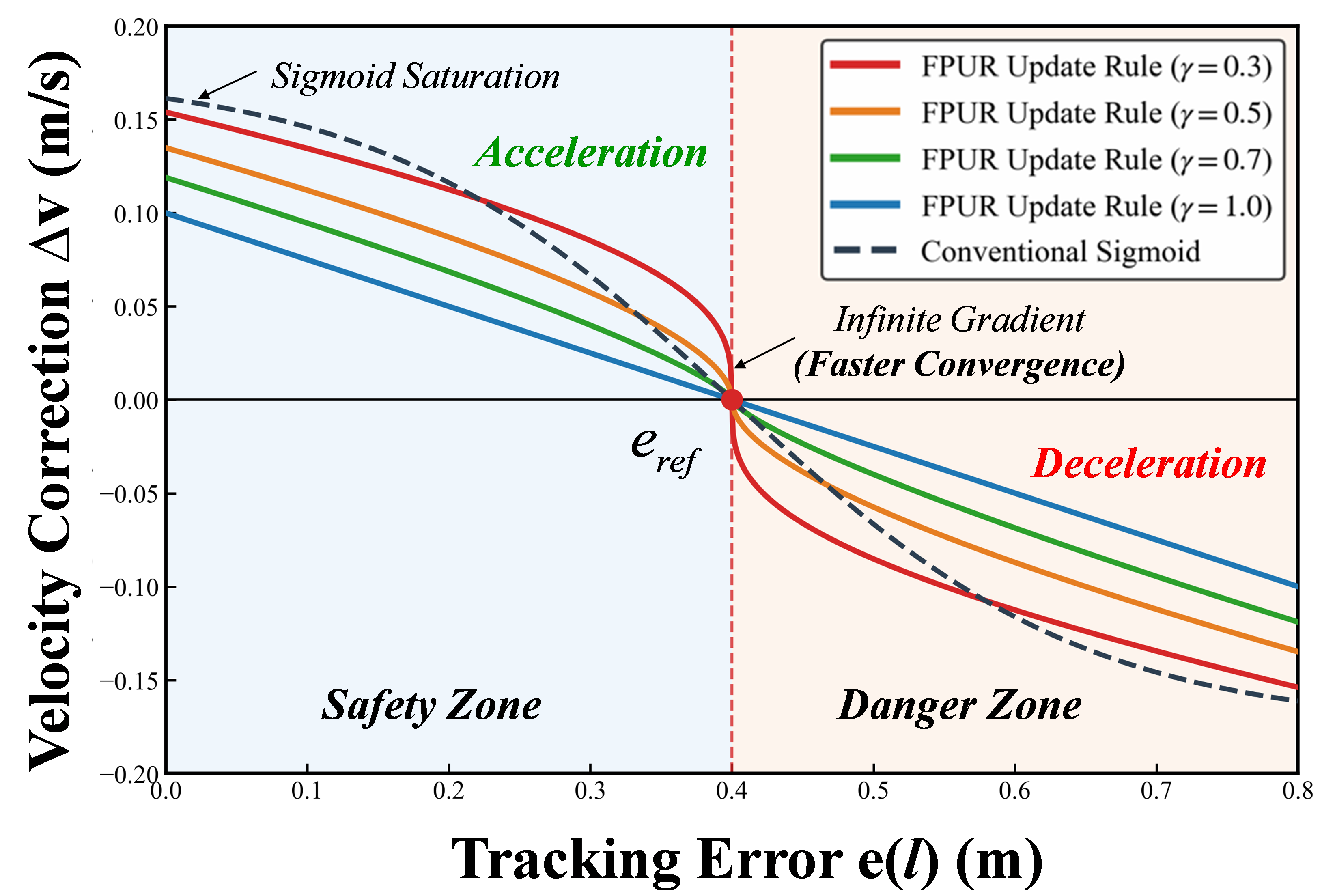}
\caption{FPUR retains stronger local correction than the saturating sigmoid update.}
\label{fig:fpur_compare}
\end{minipage}\hfill
\begin{minipage}[t]{0.48\textwidth}
\vspace{0pt}
\centering
\includegraphics[width=0.95\linewidth]{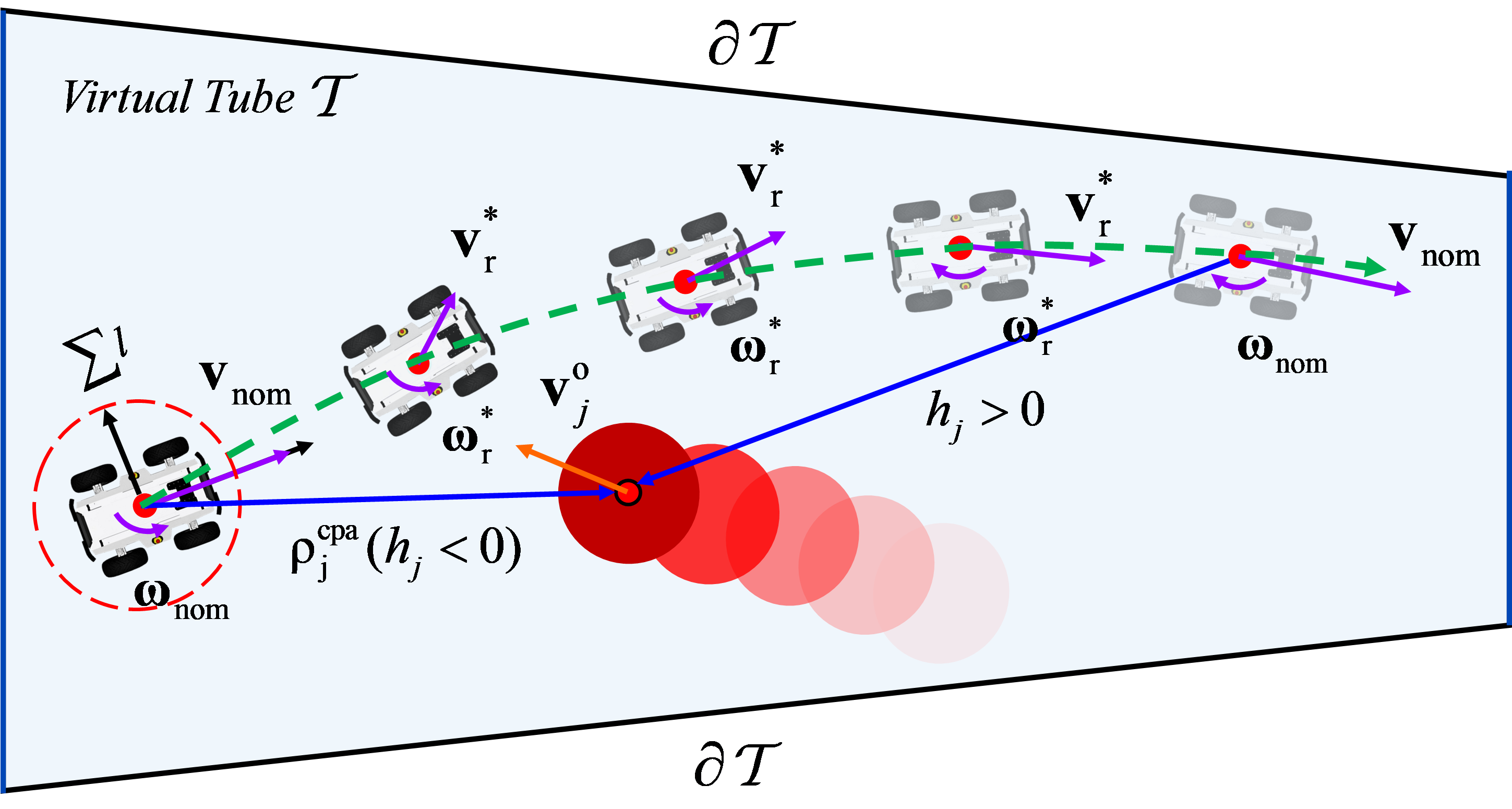}
\caption{ARB-CBF local avoidance with risk blended speed capping and tangential steering.}
\label{fig:arb_cbf}
\end{minipage}
\end{figure*}

\subsubsection{Risk-Aware MIMO Time-Optimal ILP}

To balance traversal time and path convergence under bounded control
authority, the objective function is defined as
\begin{equation}
\label{eq:pareto_functional}
J_k
=
\int_{0}^{L}
\left(
\frac{1}{v_{h,k}(l)}
+k_E V_b(e_k(l))
\right)\mathrm{d}l,
\end{equation}
where $k_E>0$ weights the tracking penalty. Increasing $v_{h,k}$
reduces traversal time, whereas $V_b$ penalizes tracking error.

Based on the Pareto analysis in our previous work~\cite{lv2026timeoptimal}, this
tradeoff is converted into an explicit spatial learning decision rather
than numerically minimizing~\eqref{eq:pareto_functional}. Choose
$0<e_0<k_a$ and define $\tilde e_k(l)=|e_k(l)|-e_0$. Thus,
$\tilde e_k(l)\leq0$ permits acceleration, whereas $\tilde e_k(l)>0$
prioritizes convergence through deceleration.

As illustrated in Fig.~\ref{fig:racing_vf}, the proposed multi-input
multi-output (MIMO) IL law jointly learns traversal-speed and
steering-bias profiles from the spatial tracking-error profile $e_k(l)$
recorded at the $N$ arc-length waypoints. Compared with the previous
single-channel ILP~\cite{lv2025high}, the MIMO formulation adds a
steering-bias learning channel for directional correction. The first-order
differential-drive kinematics in~\eqref{eq:diff_drive} are used without
requiring an inverse model or model derivatives. The MIMO IL law therefore
approaches a time-optimal Pareto profile within the tube constraint.

To incorporate the dynamic risk defined in Section~II-C, let
$\mathbf{R}_k(l)=\operatorname{diag}(1,1-W_k(l))$ and
$\mathbf{e}_v=[1,0]^\top$. The risk-aware MIMO update is
\begin{equation}
\label{eq:mimo_update}
\mathbf{z}_{k+1}(l)
=
\Pi_{\mathcal{Z}}
\left[
\mathbf{z}_k(l)
+\mathbf{K}\mathbf{R}_k(l)\bm{\Phi}_k(l)
-\mu_r W_k(l)\mathbf{e}_v
\right].
\end{equation}
Here, $\mathbf{z}_k(l)=[v_{h,k}(l),\omega_{h,k}(l)]^\top$ is the learned
profile. The correction vector is $\bm{\Phi}_k(l)=[\Phi_v(\tilde
e_k(l)),\Phi_{\omega}(e_k(l))]^\top$. The gain matrix is
$\mathbf{K}=\operatorname{diag}(-\mu_v,\mu_{\omega})$, where
$\mu_v,\mu_{\omega},\mu_r>0$ are learning gains. The projection
$\Pi_{\mathcal{Z}}(\cdot)$ enforces $\mathbf{z}_k(l)\in\mathcal{Z}$ for all
$k$ and $l$, with $\mathcal{Z}=[v_{\min},v_{\max}]\times[-\omega_{h,\max},
\omega_{h,\max}]$. The risk term suppresses aggressive speed adaptation
and attenuates transient steering learning in high-risk regions. When
$W_k(l)=0$, $\mathbf{R}_k(l)=\mathbf{I}$ and~\eqref{eq:mimo_update} reduces to
the nominal MIMO-FPUR update.
\subsubsection{FPUR-Based Learning Law}

To improve traversal performance within a limited number of learning
iterations, the IL law should provide effective correction during each
rollout. The sigmoid learning law used in our previous
ILP~\cite{lv2025high} tends to saturate, which limits its local correction
capability. Inspired by the FPUR learning law in~\cite{li2025accelerating}, we adopt this law
to strengthen finite-iteration updates. The two correction channels are defined as
\begingroup
\setlength{\abovedisplayskip}{0pt}
\setlength{\abovedisplayshortskip}{0pt}
\begin{subequations}
\label{eq:fpur}
\begin{align}
\Phi_v(\tilde e)
&=
\alpha_v\tilde e
+\beta_v|\tilde e|^{\gamma_v}\operatorname{sgn}(\tilde e),
\label{eq:fpur_speed}
\\
\Phi_{\omega}(e)
&=
\alpha_{\omega} e
+\beta_{\omega}|e|^{\gamma_{\omega}}\operatorname{sgn}(e),
\label{eq:fpur_steering}
\end{align}
\end{subequations}
\endgroup
where $\alpha_{v,\omega},\beta_{v,\omega}>0$ and
$0<\gamma_{v,\omega}<1$. The proportional term maintains effective
correction for large deviations, whereas the fractional-power term
strengthens the update near the switching threshold. For standard FPUR,
its convergence and convergence-rate properties have been established
in~\cite{li2025accelerating}. As illustrated in Fig.~\ref{fig:fpur_compare},
FPUR provides stronger local correction than the saturating sigmoid
update. The projection $\Pi_{\mathcal{Z}}$ keeps the learned profiles
within the admissible set $\mathcal{Z}$.

After $K$ iterations, the learned profiles
$\{v_h^\ast,\omega_h^\ast\}$ are stored by arc length. At runtime, the closest
waypoint $i^\ast$ generates
\begin{equation}
\label{eq:nominal_command}
v_{\mathrm{nom}}=v_h^\ast(l_{i^\ast}), \qquad
\omega_{\mathrm{nom}}
=
k_\theta e_\theta+\omega_h^\ast(l_{i^\ast}),
\end{equation}
which serves as the nominal input to the safety filter in
Section~III-C.
\subsection{Anticipatory Risk Blended Control Barrier Function}
\label{sec:cbf_filter}

Based on the time-efficient nominal command learned in the previous
section, this section further provides runtime safety correction for
dynamic obstacle avoidance. As shown in Fig.~\ref{fig:arb_cbf}, the proposed
Anticipatory Risk Blended Control Barrier Function (ARB-CBF) converts
the nominal command into a safe command using the risk information
defined in Section~II-C. The hard CBF caps the forward speed, while
$W_{\mathrm{risk}}$ adjusts the speed and steering for anticipatory
avoidance.

Let $\mathcal O$ denote the detected obstacles and $\mathcal J_{\rm h}=\{j\in\mathcal O\mid\cos\phi_j>0\}$ the hard-constraint set in the forward sensing sector. For obstacle $j$, define $X_j=[\mathbf x^\top,(\mathbf p_j^{\rm o})^\top,(\mathbf{v}_j^{\rm o})^\top]^\top$. Following CBF theory~\cite{zeng2021safety}, the single obstacle and joint safe sets and the associated CBF condition, with $\gamma$ an extended class-$\mathcal K_\infty$ function, are formulated as
\begin{subequations}\label{eq:cbf_safe_condition}
\begin{gather}
\mathcal C_j=\{X_j\mid h_j(X_j)\geq0\},\quad
\mathcal C=\bigcap_{j\in\mathcal J_{\rm h}}\mathcal C_j,
\label{eq:safe_set}\\
\sup_{\mathbf u_r\in\mathcal U_r}\!\left[\dot h_j(X_j,\mathbf u_r)\right]\geq-\gamma\!\left(h_j(X_j)\right),\quad \gamma\in\mathcal K_{\infty}^{e}.
\label{eq:cbf_condition}
\end{gather}
\end{subequations}

For circular clearance, the hard barrier is $h_j=\rho_j^2-R_{0,j}^2$, where $h_j>0$, $h_j=0$, and $h_j<0$ correspond to safe clearance, the hard boundary, and violation of the protected region, respectively. Differentiating $h_j$ using the relative-distance dynamics in~(\ref{eq:rho_dot}) yields
\begin{equation}
\dot h_j=2\rho_j\big(-v_r\cos\phi_j+\mathbf b_j^\top\mathbf{v}_j^{\rm o}\big).
\label{eq:hdot}
\end{equation}

Because $\omega_r$ is absent from~\eqref{eq:hdot}, the KKT conditions decouple, reducing the minimum-intervention correction to a scalar speed cap. The safe speed and global hard CBF cap are therefore
\begin{subequations}\label{eq:hard_cap}
\begin{gather}
v_j^{\rm safe}=
\frac{\gamma(h_j)+2\rho_j\mathbf b_j^\top\mathbf{v}_j^{\rm o}}{2\rho_j\cos\phi_j},
\label{eq:safe_vel}\\
\bar{v}=
\begin{cases}
\min\!\left(v_{\max},\min_{j\in\mathcal J_{\rm h}}v_j^{\rm safe}\right),&\mathcal J_{\rm h}\neq\varnothing,\\
v_{\max},&\mathcal J_{\rm h}=\varnothing.
\end{cases}
\label{eq:global_safe}
\end{gather}
\end{subequations}
Thus, the angular command is not involved in the hard CBF projection, and the online computation only requires a linear scan over the detected obstacles.

For anticipatory tangential avoidance, the dominant risk obstacle is selected as $j^*=\arg\max_{j\in\mathcal J_{\rm r}}\lambda_j$. At first activation, a steering side $\sigma\in\{-1,1\}$ is latched until $\mathcal J_{\rm r}$ becomes empty, yielding $\phi^\perp=\operatorname{wrap}(\phi_{j^*}+\sigma\pi/2)$.

The anticipatory and hard safety layers are combined at the controller output as
\begin{subequations}\label{eq:blended_cmd}
\begin{gather}
v_r^*=\min\!\left\{\bar{v},\,v_{\rm nom}-W_{\rm risk}(v_{\rm nom}-v_{\min})\right\},\\
\omega_r^*=\omega_{\rm nom}+W_{\rm risk}(k_{\omega}\phi^\perp-\omega_{\rm nom}).
\end{gather}
\end{subequations}
When $W_{\mathrm{risk}}=0$, the nominal command is retained subject to the hard CBF speed cap. As risk increases, the command is smoothly adjusted toward safer speed and steering. Since $v_r^\ast\leq\bar{v}\leq v_j^{\mathrm{safe}}$ for all $j\in\mathcal{J}_h$, \eqref{eq:hdot} gives $\dot h_j\geq-\gamma(h_j)$. Thus, under feasible CBF constraints, the joint hard-safe set remains forward invariant.

\begin{remark}
The ARB-CBF formulation offers three advantages. First, its local polar
representation requires onboard LiDAR and IMU, while
$R_{0,j}$ and $R_{\mathrm{safe},j}$ distinguish the hard safety boundary
from the anticipatory response region. Second, because $\omega_r$ is
absent from (22), the KKT conditions yield the explicit scalar speed cap
in (23a)--(23b), which is evaluated with linear computational complexity.
Unlike optimization-based CBF and MPC-CBF methods~\cite{liang2025pointcloud,jian2023dynamic,zhang2026extended},
it requires no online QP solution. Third, risk
blending and latched steering provide a smooth transition between nominal
tracking and avoidance while preserving the hard CBF constraint.
\end{remark}

\subsection{Integrated ILP-CBF Solution Framework}

MIMO-FPUR-ILP in Section~III-B addresses $\mathbf{P}_1$ by learning
bounded traversal-speed and steering-bias profiles for time-optimal
nominal motion, while ARB-CBF in Section~III-C addresses $\mathbf{P}_2$
by correcting the executed command to satisfy dynamic-obstacle safety
constraints. The learned profiles generate $\mathbf{u}_{\mathrm{nom}}$, which
serves as the nominal input to ARB-CBF. In this way, $\mathbf{P}_1$
maintains time-efficient nominal motion, whereas $\mathbf{P}_2$ enforces
runtime safety. Their coordinated solution addresses Problem~$\mathbf{P}$
without repeatedly solving the coupled optimization online, as summarized
in Algorithm~1.

The computational cost consists of ILP learning and runtime safety
filtering. Since the ILP update is performed waypoint-wise, $k$ learning
iterations over $N$ path waypoints lead to $O(kN)$ complexity.
ARB-CBF evaluates obstacle safety quantities by scanning perceived obstacles, resulting in linear computational complexity.In contrast,
optimization-based safety-critical methods typically require $O(n^2)$
computational complexity with respect to the optimization dimension
$n$. Therefore, the proposed framework avoids repeated online
optimization and retains lightweight runtime computation.

\begin{algorithm}[t]
\caption{Safety-Critical ILP-CBF Planner}
\label{alg:ilp_cbf}
\begin{algorithmic}[1]
\REQUIRE Reference path $\mathcal P$, initial profiles
$\{v_{h,0},\omega_{h,0}\}$, state $\mathbf{x}$, and obstacles $\mathcal O$
\ENSURE Safe command $\mathbf{u}_r=[v_r,\omega_r]^\top$
\STATE Set $W_k(l)=0$ and learn nominal profiles by (18)--(19)
\STATE Obtain $\mathbf{u}_{\mathrm{nom}}$
$=[v_{\mathrm{nom}},\omega_{\mathrm{nom}}]^\top$
from the closest waypoint $i^*$ by (20)
\STATE Estimate obstacle states and $W_{\mathrm{risk}}$ (Section~II-C)
\STATE \textbf{if} replanning is triggered \textbf{then}
\STATE \hspace{\algorithmicindent} Map $W_{\mathrm{risk}}(t)$ to $W_k(l)$, update profiles, refresh $\mathbf{u}_{\mathrm{nom}}$
\STATE \textbf{end if}
\STATE Compute $\mathcal J_h$, $v_j^{\mathrm{safe}}$, and $\bar{v}$ by (23)
\STATE Determine $j^*$ and $\sigma$ when $W_{\mathrm{risk}}>0$, and compute
$\mathbf{u}_r^*=[v_r^*,\omega_r^*]^\top$ by (24)
\STATE \textbf{if} CBF constraint is infeasible \textbf{then}
\STATE \hspace{\algorithmicindent} $\mathbf{u}_r\leftarrow\mathbf{0}$; trigger emergency replanning/recovery
\STATE \textbf{else}
\STATE \hspace{\algorithmicindent} $\mathbf{u}_r\leftarrow\mathbf{u}_r^*$
\STATE \textbf{end if}
\RETURN $\mathbf{u}_r$
\end{algorithmic}
\end{algorithm}

\begin{remark}
The integrated ILP-CBF architecture follows separation and minimal-intervention principles. The learned nominal behavior is preserved during normal execution, while ARB-CBF modifies only the executed command under safety constraints. Moreover, risk blending maintains the forward command above $v_{\min}$; whenever the hard CBF constraint admits a speed no smaller than $v_{\min}>0$, the final command remains positive. Therefore, under strict positive feasibility, the safety filter cannot induce a stationary deadlock.
\end{remark}
\section{Simulations and Experiments}
\label{sec:experiments}

This section presents comprehensive simulations and real-world experiments\footnote{\url{https://b23.tv/BV1MDeu6fE36}} to evaluate the performance and computational efficiency of the proposed algorithm\footnote{\url{https://github.com/WLM-boop/D-ILP}}. First, comparative evaluations on learning-law convergence, alongside static and dynamic map benchmarks against representative open-source planners, are conducted in the lightweight IR-SIM environment~\cite{han2026ir}. Second, a high-fidelity corridor scenario with moving pedestrians is implemented in Gazebo to evaluate dynamic safety. Finally, onboard experiments on an AgileX LIMO Pro mobile platform validate real-time obstacle avoidance under unmapped indoor environments. Unless otherwise specified, all simulation evaluations were conducted on a desktop workstation featuring an Intel Core i7-14700 CPU and 64~GB RAM.

\subsection{Stability and Performance Comparison of IL Laws}
\label{sec:performance_comparison}

\begin{figure}[!t]
\centering
\includegraphics[width=0.48\columnwidth]{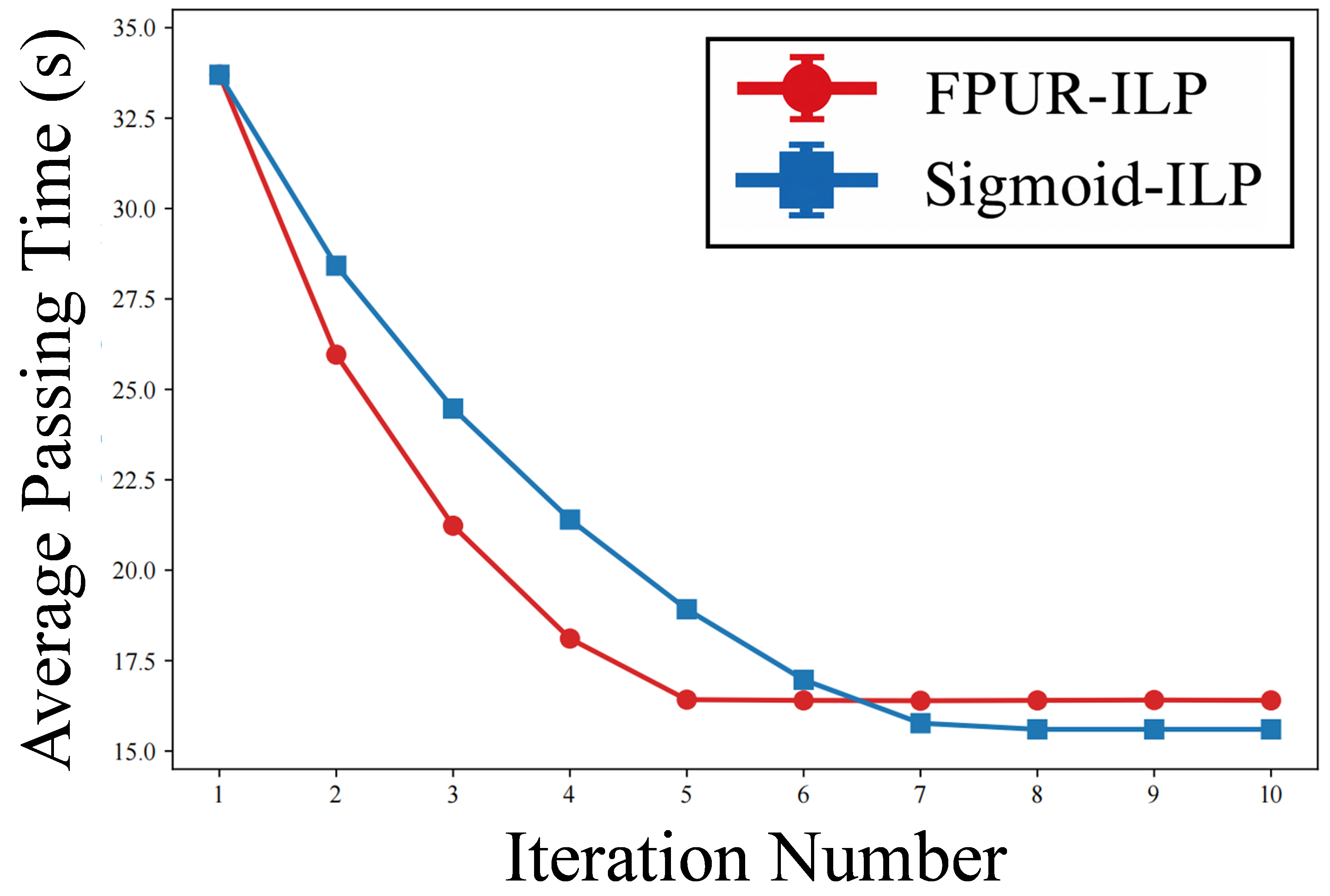}\hfill
\includegraphics[width=0.48\columnwidth]{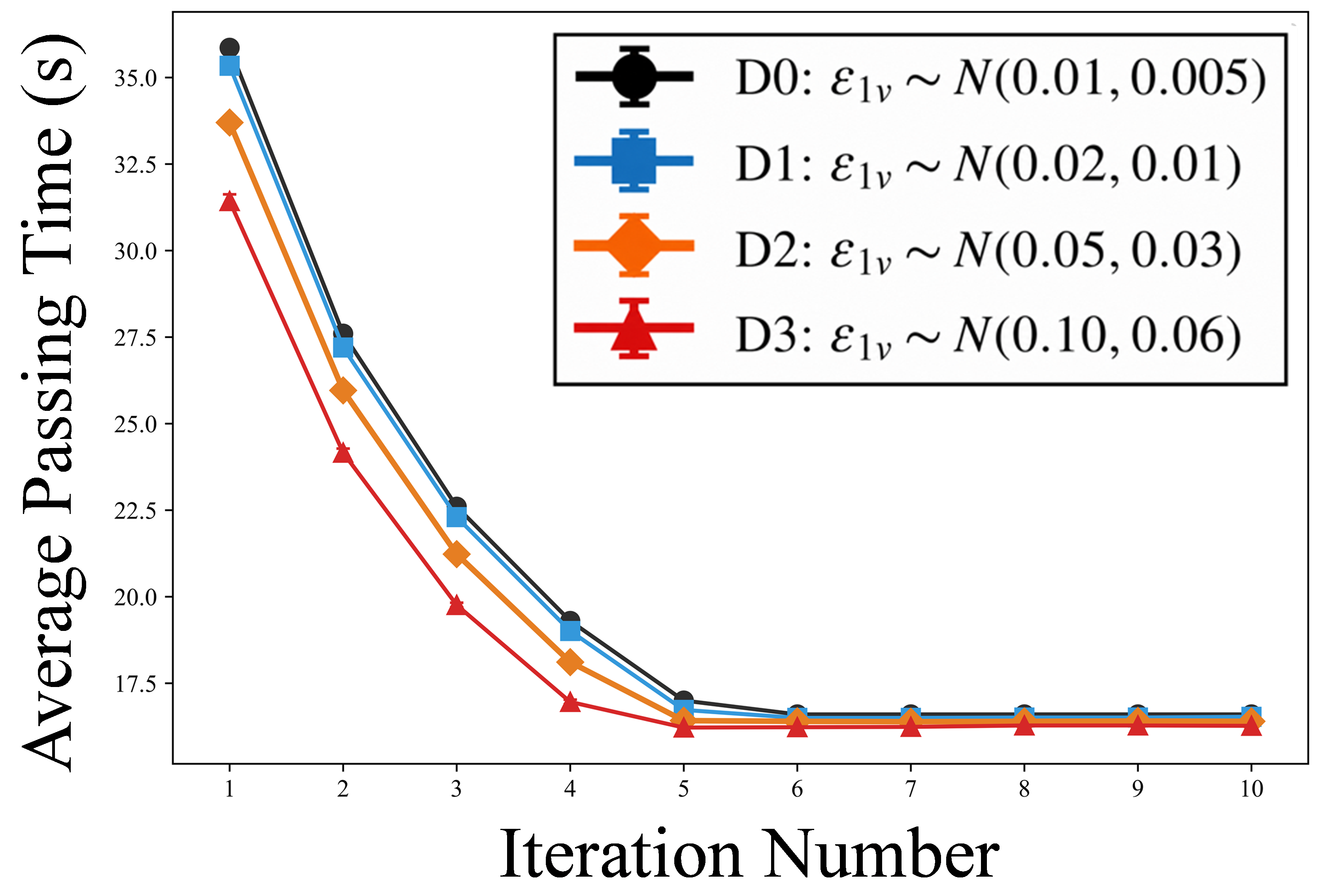}\\[0.5mm]
\makebox[0.48\columnwidth][c]{\footnotesize(a) S curve convergence}\hfill
\makebox[0.48\columnwidth][c]{\footnotesize(b) Learning stability}\\[1.5mm]
\includegraphics[width=0.48\columnwidth]{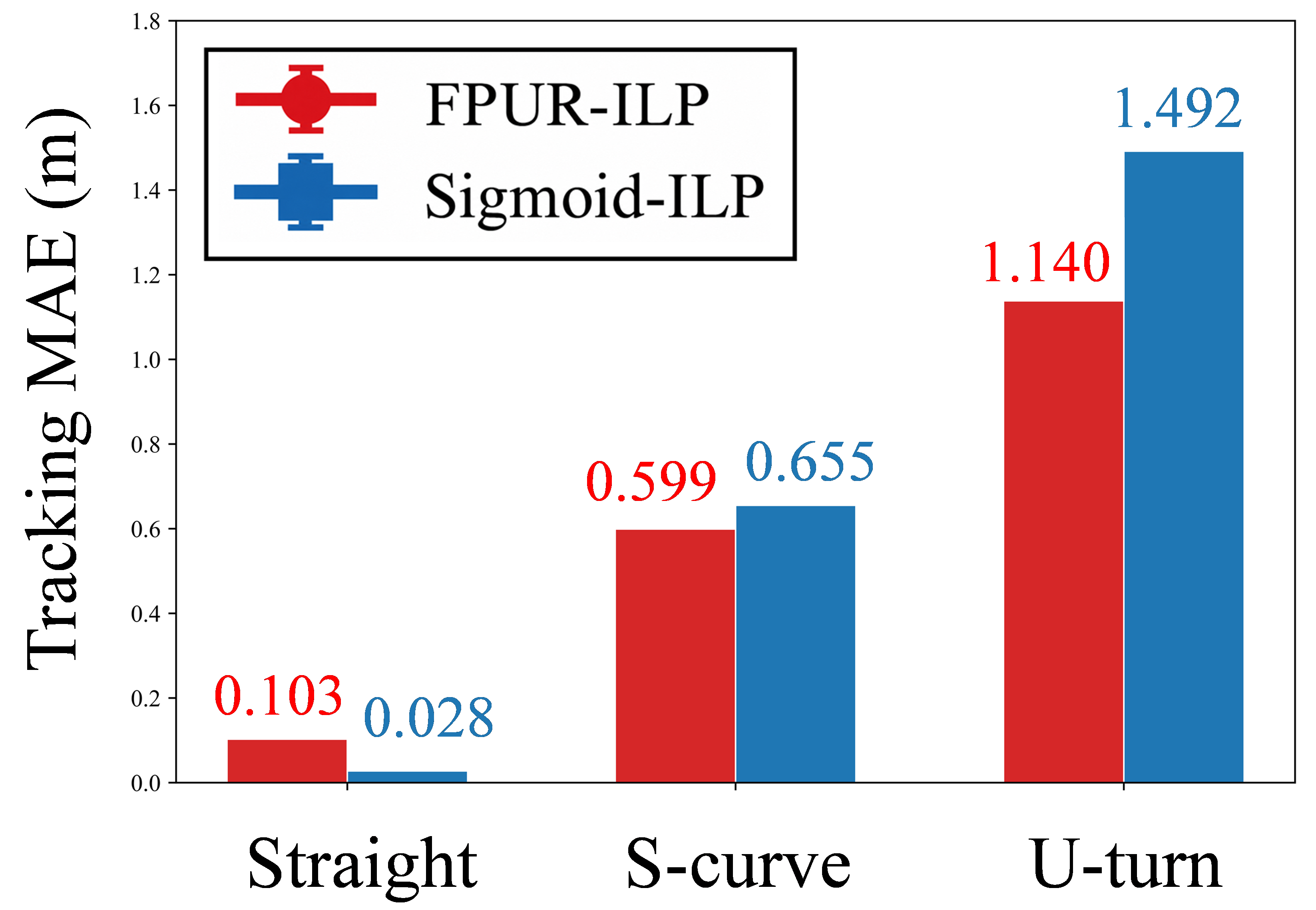}\hfill
\includegraphics[width=0.48\columnwidth]{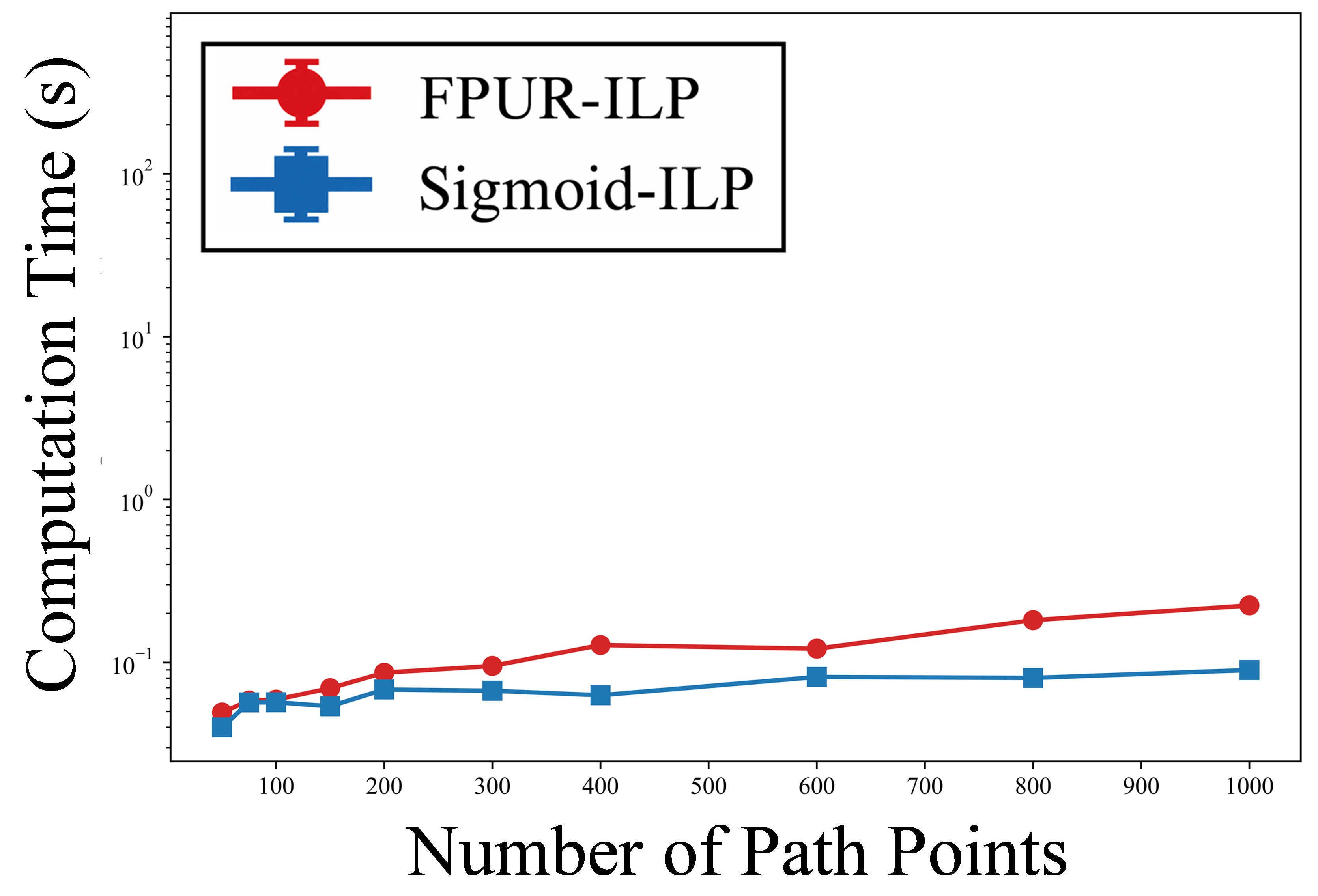}\\[0.5mm]
\makebox[0.48\columnwidth][c]{\footnotesize(c) Tracking MAE}\hfill
\makebox[0.48\columnwidth][c]{\footnotesize(d) Computation time}
\caption{Comparison of IL Laws Under Given Paths.}
\label{fig:mimo_fpur_results}
\end{figure}

This subsection compares the proposed FPUR-ILP against the Sigmoid-ILP baseline from our prior work~\cite{lv2025high} across fixed straight, S-curve, and U-turn path geometries. For each path, 10 Monte Carlo trials are executed in IR-SIM for up to 10 learning iterations, with a practical learning budget capped at five iterations. Gaussian disturbances of varying intensities are injected to model environmental perturbations. Detailed performance comparisons are illustrated in Fig.~\ref{fig:mimo_fpur_results} and summarized in Table~\ref{tab:performance_d2}.

\begin{table}[!t]
\setlength{\textfloatsep}{0pt}
\caption{Performance Comparison of IL Laws Under Given Paths.}
\label{tab:performance_d2}
\centering
\setlength{\tabcolsep}{2.2pt}
\renewcommand{\arraystretch}{1.08}
\resizebox{\columnwidth}{!}{%
\begin{tabular}{@{}lcccccc@{}}
\toprule
& \multicolumn{2}{c}{$T_5$ (s)} & \multicolumn{2}{c}{$T_{10}$ (s)} & \multicolumn{2}{c}{MAE (m)} \\
\cmidrule(lr){2-3}\cmidrule(lr){4-5}\cmidrule(l){6-7}
Path & FPUR-ILP & Sigmoid-ILP & FPUR-ILP & Sigmoid-ILP & FPUR-ILP & Sigmoid-ILP \\
\midrule
Straight & \textbf{8.9} & 10.1 & 8.9 & \textbf{8.6} & 0.103 & \textbf{0.028} \\
S curve  & \textbf{16.4} & 18.9 & 16.4 & \textbf{15.6} & \textbf{0.599} & 0.655 \\
U turn   & \textbf{16.3} & 19.2 & \textbf{15.9} & 16.7 & \textbf{1.140} & 1.492 \\
\bottomrule
\end{tabular}}
\end{table}

As illustrated in Fig.~\ref{fig:mimo_fpur_results}(a) and Table~\ref{tab:performance_d2}, FPUR-ILP consistently yields the lowest \(T_{5}\) across all evaluated path geometries. Specifically, on the S-curve path, \(T_{5}\) drops to \(16.4~\mathrm{s}\), outperforming Sigmoid-ILP (\(18.9~\mathrm{s}\)). Although Sigmoid-ILP records a lower \(T_{10}\) on the straight and S-curve tracks, this convergence occurs beyond the target iteration budget. Furthermore, Fig.~\ref{fig:mimo_fpur_results}(b) confirms that FPUR-ILP stabilizes within a narrow passing-time envelope after approximately five iterations under environmental perturbations. Fig.~\ref{fig:mimo_fpur_results}(c) and Table~\ref{tab:performance_d2} highlight lower MAEs for FPUR-ILP on the S-curve and U-turn paths, while Sigmoid-ILP exhibits superior tracking accuracy on the straight segment. As depicted in Fig.~\ref{fig:mimo_fpur_results}(d), FPUR-ILP introduces a slight computational overhead owing to its joint MIMO profile update. However, both learning mechanisms execute well within \(0.15~\mathrm{s}\) for paths containing up to 600 waypoints. In summary, FPUR-ILP achieves rapid finite-iteration convergence and robust learning stability while maintaining a lightweight computational footprint.

\subsection{Benchmarking Against Local Planning Baselines}
\label{sec:baseline_comparison}

\begin{table*}[!t]
\caption{Comparison Results in Static and Dynamic Random Maps in IR-SIM.}
\label{tab:baseline_comparison}
\centering
\setlength{\tabcolsep}{2.2pt}
\renewcommand{\arraystretch}{1.0}
\resizebox{\textwidth}{!}{%
\begin{tabular}{lccccccc|ccccccc}
\toprule
& \multicolumn{7}{c|}{Static Map} & \multicolumn{7}{c}{Dynamic Map} \\
\cmidrule(lr){2-8}\cmidrule(lr){9-15}
Method & SR (\%) & PT (s) & AS (m/s) & AC (rad/m) & APT (ms) & MSM (m) & ASM (m) & SR (\%) & PT (s) & AS (m/s) & AC (rad/m) & APT (ms) & MSM (m) & ASM (m) \\
\midrule
MCBF & \textbf{100} & 19.08 & 2.47 & 0.453 & 2.585 & \textbf{2.14} & 3.51
& \textbf{100} & 18.71 & 2.46 & 0.451 & 1.139 & 1.50 & 1.67 \\
TEB & 90 & 14.32 & 3.29 & 0.162 & 8.415 & 1.64 & 3.43
& \textbf{100} & 14.26 & 3.12 & 0.120 & 5.965 & 1.38 & 1.54 \\
CSC-MPPI & \textbf{100} & 10.80 & 4.16 & 0.141 & 9.781 & 1.73 & 3.54
& \textbf{100} & 13.64 & 3.24 & \textbf{0.107} & 8.822 & 1.21 & \textbf{2.84} \\
ILP & \textbf{100} & \textbf{10.02} & \textbf{4.84} & \textbf{0.124} & \textbf{0.109} & 1.31 & 3.37
& 85 & 13.15 & 3.21 & 0.135 & 0.431 & 1.08 & 1.26 \\
Proposed & \textbf{100} & 10.43 & 4.58 & 0.132 & 0.122 & 1.81 & \textbf{3.57}
& \textbf{100} & \textbf{12.85} & \textbf{3.63} & 0.113 & \textbf{0.188} & \textbf{2.29} & 2.59 \\\bottomrule
\end{tabular}}

\begin{minipage}{0.96\textwidth}
\footnotesize SR: success rate, PT: passing time, AS: average speed, AC: average curvature, APT: average planning time, MSM: minimum safety margin, ASM: average safety margin.
\end{minipage}
\end{table*}

\begin{figure*}[!t]
\centering
\setlength{\abovecaptionskip}{2pt}
\setlength{\tabcolsep}{0pt}
\begin{tabular}{@{}cccccc@{}}
\includegraphics[height=0.18\textwidth]{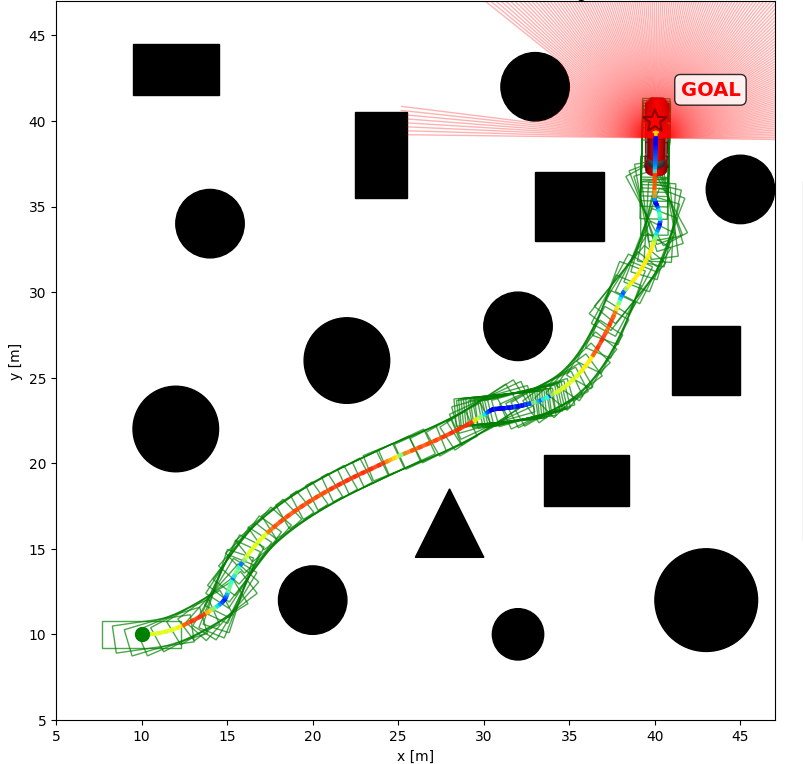} &
\includegraphics[height=0.18\textwidth]{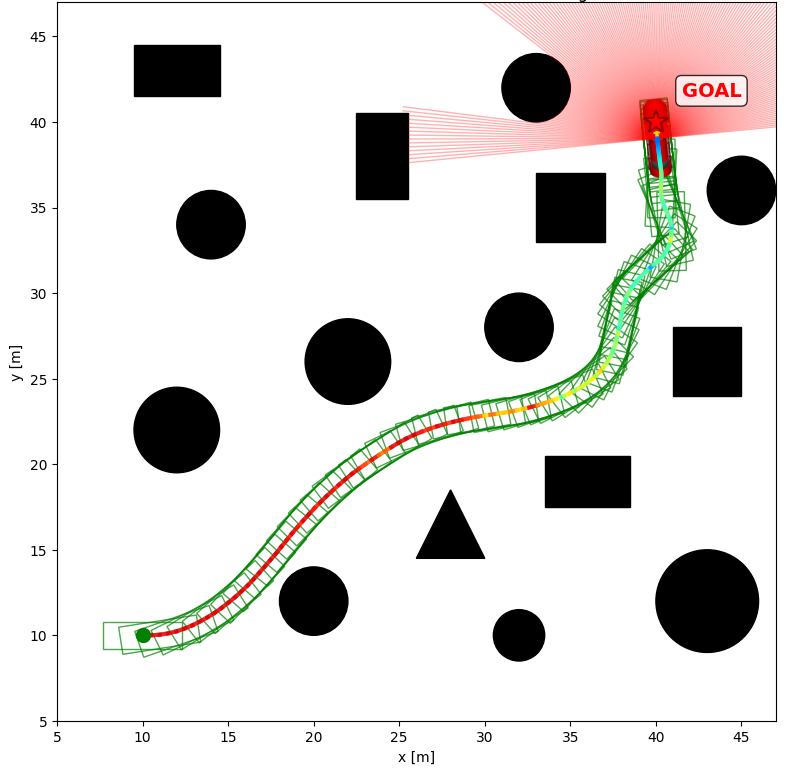} &
\includegraphics[height=0.18\textwidth]{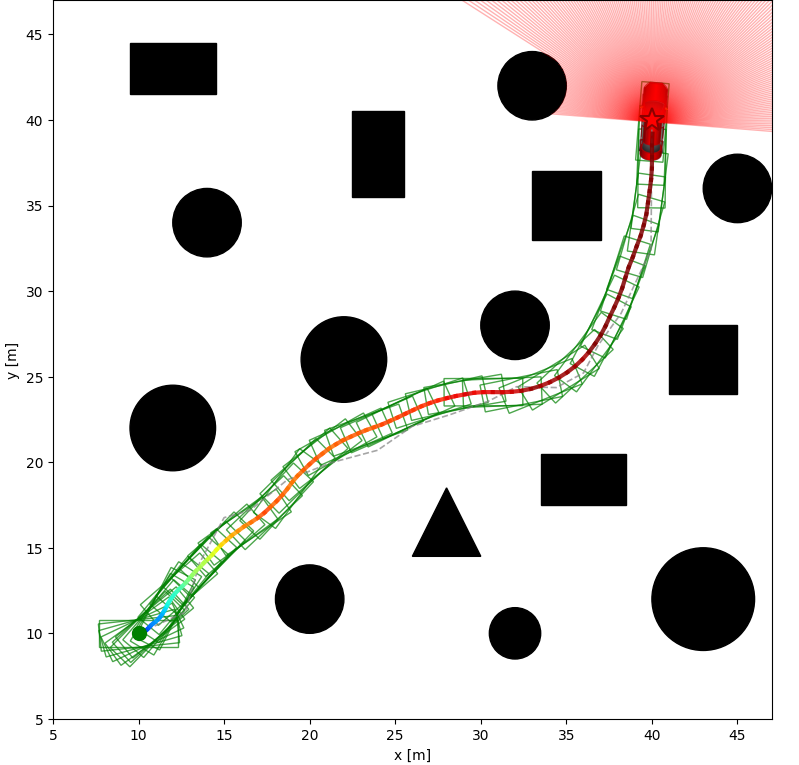} &
\includegraphics[height=0.18\textwidth]{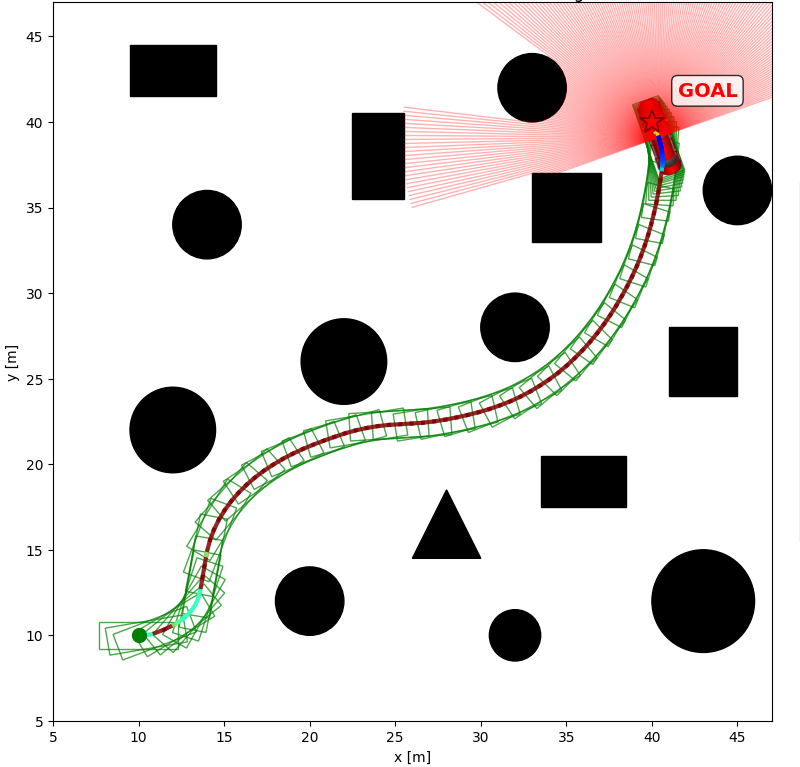} &
\includegraphics[height=0.18\textwidth]{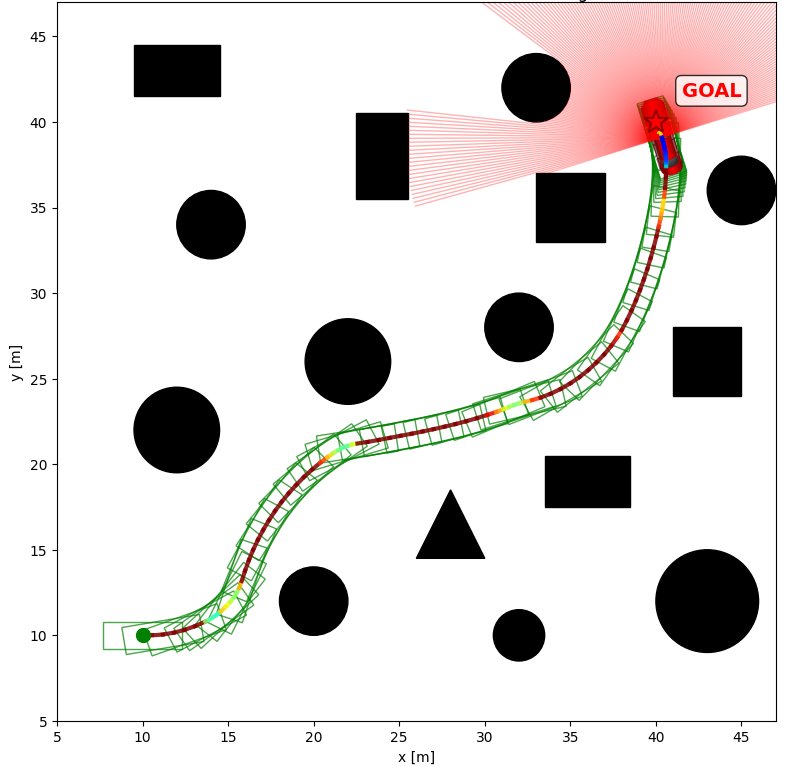} &
\includegraphics[height=0.18\textwidth]{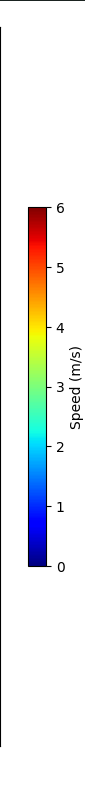} \\
{\footnotesize(a) MCBF (17.70 s)} & {\footnotesize(b) TEB (14.20 s)} & {\footnotesize(c) CSC-MPPI (10.80 s)} & {\footnotesize(d) ILP (10.10 s)} & {\footnotesize(e) Proposed (10.30 s)} & {} \\[0.8mm]
\includegraphics[height=0.18\textwidth]{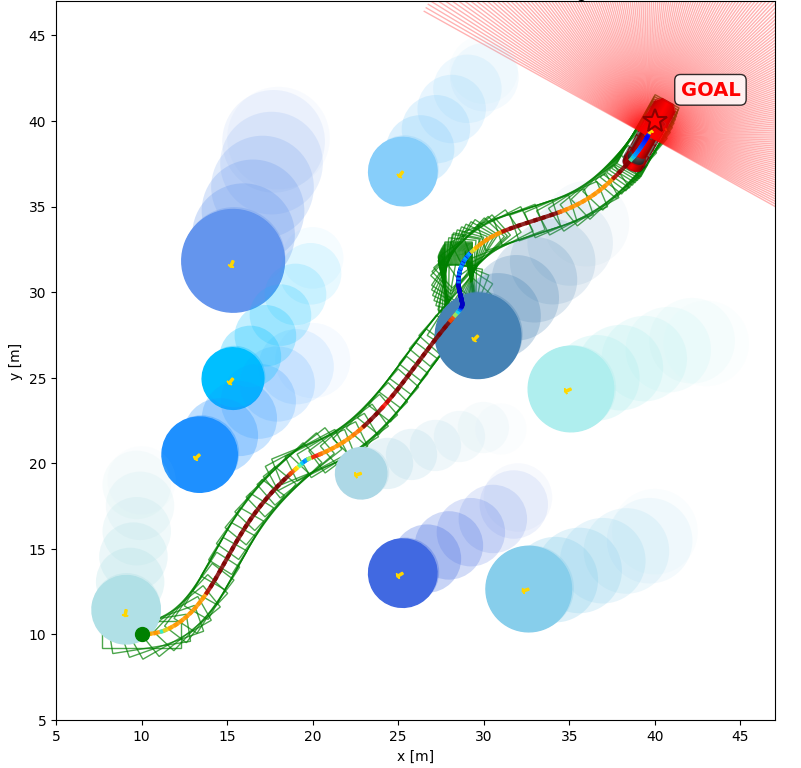} &
\includegraphics[height=0.18\textwidth]{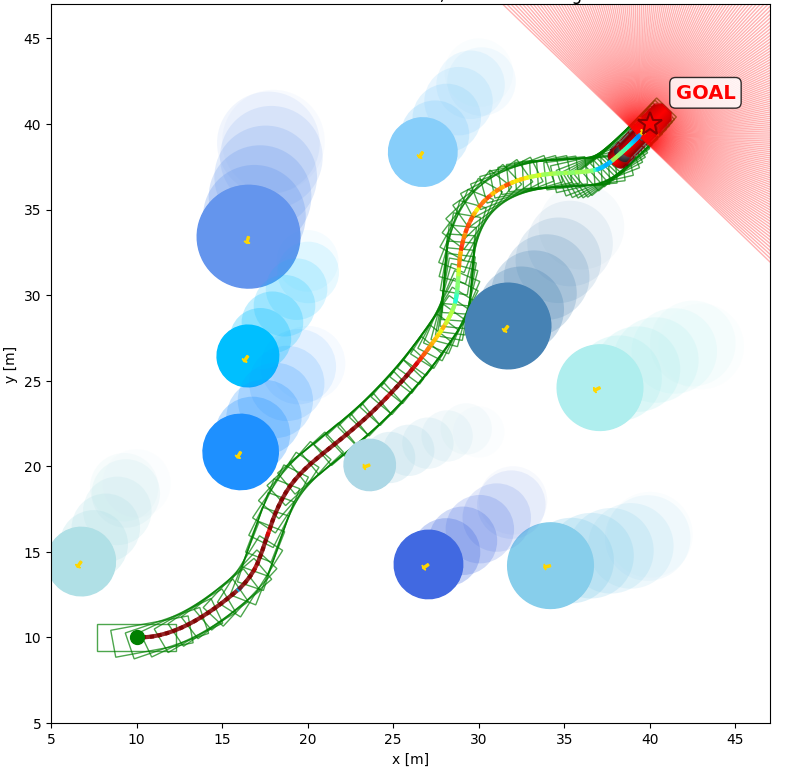} &
\includegraphics[height=0.18\textwidth]{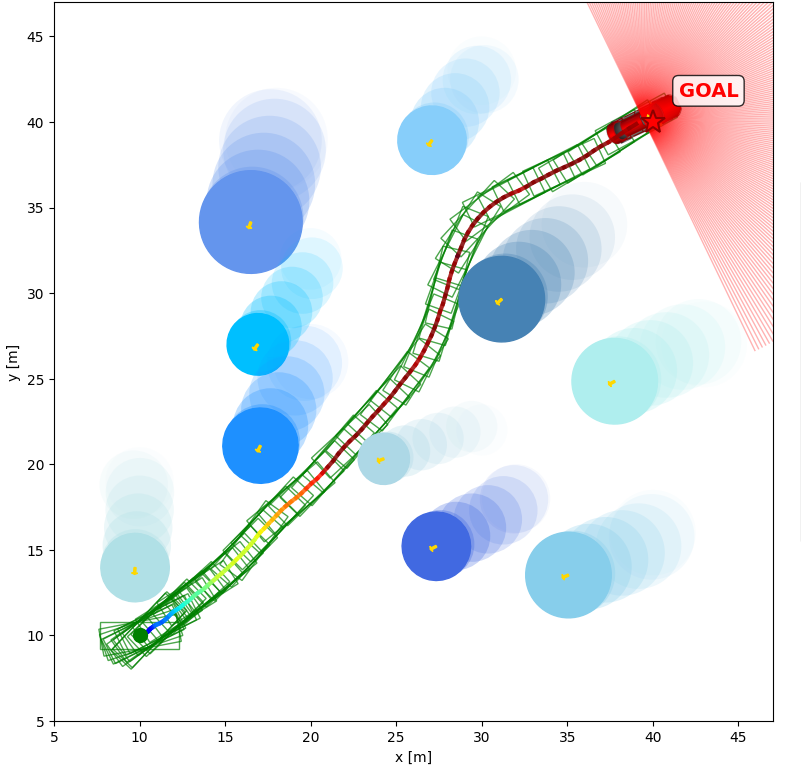} &
\includegraphics[height=0.18\textwidth]{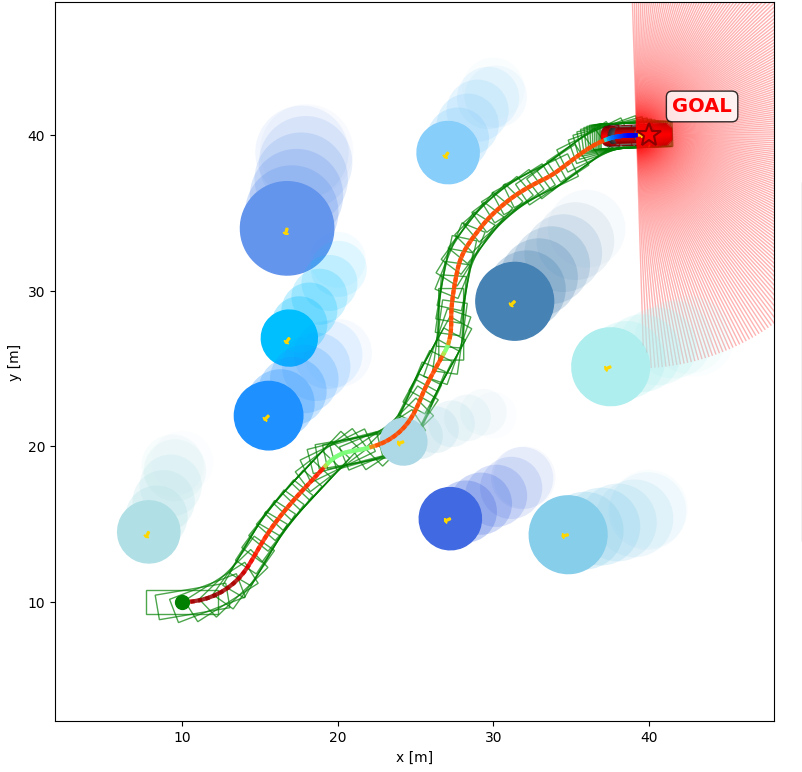} &
\includegraphics[height=0.18\textwidth]{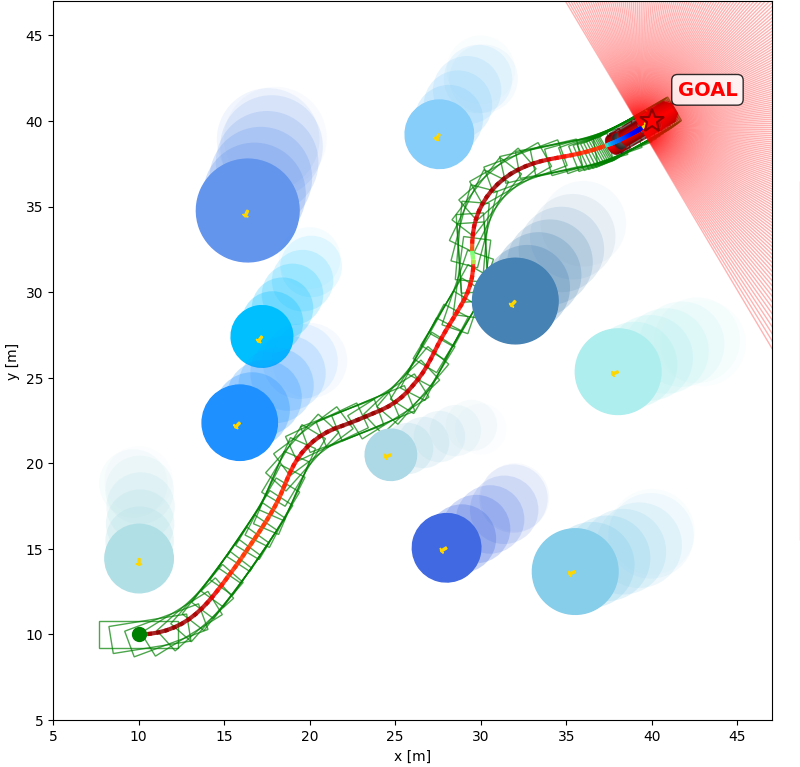} &
\includegraphics[height=0.18\textwidth]{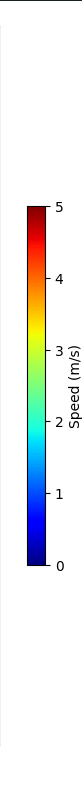} \\
{\footnotesize(f) MCBF (18.30 s)} & {\footnotesize(g) TEB (14.50 s)} & {\footnotesize(h) CSC-MPPI (13.10 s)} & {\footnotesize(i) ILP (13.20 s)} & {\footnotesize(j) Proposed (12.10 s)} & {}
\end{tabular}
\caption{Navigation process and trajectory comparison of local planners in static and dynamic IR-SIM environments.}
\label{fig:baseline_comparison}
\end{figure*}

The proposed framework was benchmarked against state-of-the-art baselines, including Modulated Control Barrier Functions (MCBF)~\cite{xue2026proactive}, Timed Elastic Band (TEB)~\cite{rosmann2017kinodynamic}, Constrained Sampling Cluster Model Predictive Path Integral (CSC-MPPI)~\cite{park2025cscmppi}, and standard ILP~\cite{lv2025high}, within the Python-based IR-SIM environment. Each planner was evaluated across 20 trials under identical start-to-goal configurations, sensor parameters, and dynamic limits. The differential-drive robot, measuring \(4.6~\text{m} \times 1.6~\text{m}\), navigated from \((10~\text{m}, 10~\text{m}, 0~\text{rad})\) to \((40~\text{m}, 40~\text{m}, 0~\text{rad})\) with \(0.1\text{s}\) control update intervals.

The static scenario featured a \(42~\text{m} \times 42~\text{m}\) environment containing 15 randomly positioned obstacles alongside a continuously updated global RRT* reference, with speed capped at \(6~\text{m/s}\). The dynamic scenario incorporated 10 moving circular obstacles (radius: \(1.5\text{--}3.0~\text{m}\)) executing random linear/angular maneuvers bounded by \([-1.5, 1.5]~\text{m/s}\) and \([-0.6, 0.6]~\text{rad/s}\), respectively, with a maximum speed of \(5~\text{m/s}\). All algorithms shared identical acceleration (\(0.8~\text{m/s}^2\)) and angular velocity (\(1.0~\text{rad/s}\)) bounds. Table~\ref{tab:baseline_comparison} summarizes the Success Rate (SR), Passing Time (PT), Average Speed (AS), Average Curvature (AC), Average Planning Time (APT), Minimum Safety Margin (MSM), and Average Safety Margin (ASM). Here, PT and AS measure temporal efficiency, whereas APT captures online computational latency per control cycle.

As summarized in Table~\ref{tab:baseline_comparison} and Fig.~\ref{fig:baseline_comparison}, the proposed method achieved a PT of \(10.43~\text{s}\) and an AS of \(4.58~\text{m/s}\) in static environments, closely matching ILP while outperforming MCBF, TEB, and CSC-MPPI. Although its PT increased by \(4.1\%\) relative to nominal ILP, the proposed framework yielded the highest ASM (\(3.57~\text{m}\)), demonstrating that the runtime safety filter enforces collision avoidance with minimal disturbance to nominal motion. Furthermore, an APT of \(0.122~\text{ms}\) confirms the ultra-lightweight execution of explicit ARB-CBF corrections without online reoptimization.

In dynamic environments, the proposed planner achieved a \(100\%\) success rate with the shortest PT (\(12.85~\text{s}\)) and highest AS (\(3.63~\text{m/s}\)). It also recorded the lowest APT and largest MSM across all baselines. Trajectory profiles in Fig.~\ref{fig:baseline_comparison} illustrate smooth progress while dynamically reacting to obstacles. In contrast, standard ILP suffered three task failures due to reliance on frequent global replanning, which elevated its APT to \(0.431~\text{ms}\). While MCBF, TEB, and CSC-MPPI achieved safe navigation, they incurred significant time delays or heavy computational overhead. Overall, the proposed method reduced dynamic APT by \(83.5\%\text{--}97.9\%\) compared to optimization-based baselines, establishing a superior balance between traversal speed, guaranteed safety, and real-time computational efficiency.

\vspace{-1pt}
\subsection{Simulations in Gazebo Simulator}
\label{sec:gazebo_simulations}

\begin{figure}[!t]
\centering

\includegraphics[width=0.98\columnwidth,trim=0 55 0 25,clip]{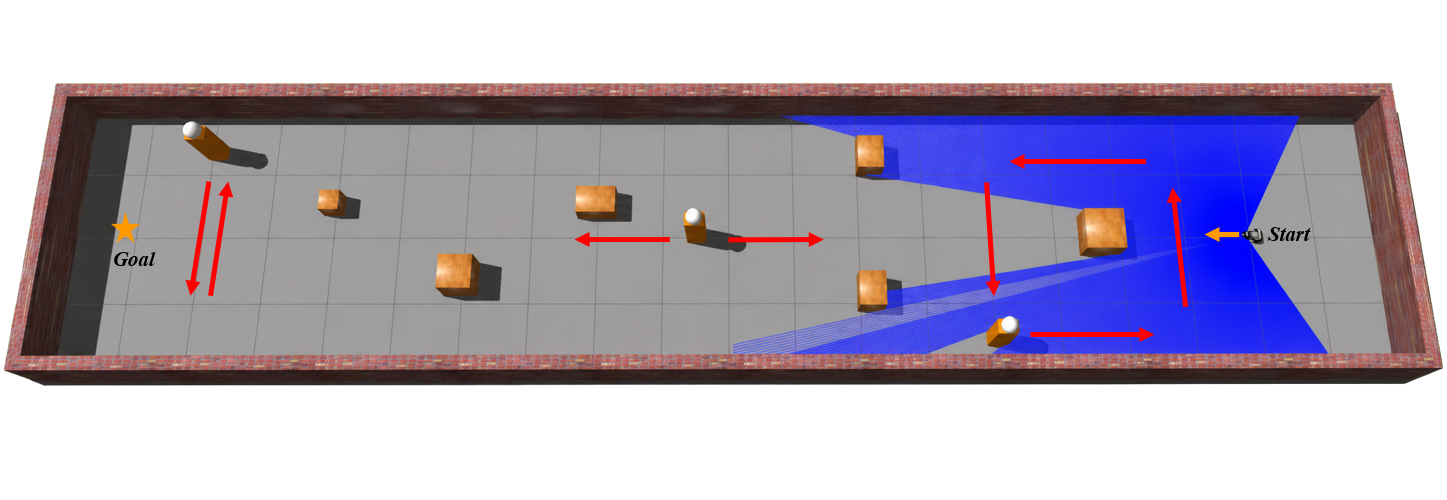}\\[0.3mm]
\makebox[\columnwidth][c]{\footnotesize(a) Gazebo based corridor simulation environment}\\[0.8mm]
\includegraphics[width=0.98\columnwidth]{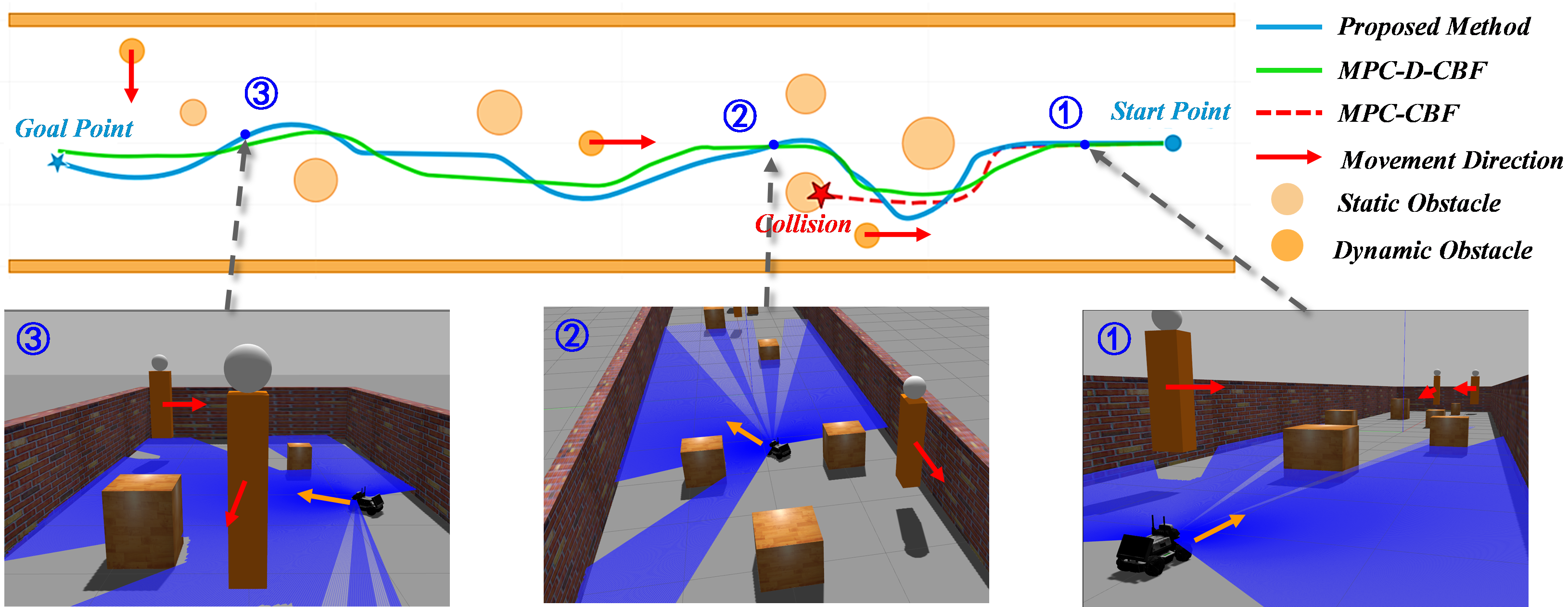}\\[0.3mm]
\makebox[\columnwidth][c]{\footnotesize(b) Qualitative comparison of the three safety-critical planners}

\caption{Gazebo corridor simulation comparison of the proposed method with MPC-CBF and MPC-D-CBF.}
\label{fig:gazebo_comparison}

\vspace{-1.2\baselineskip}
\end{figure}

\begin{table}[!t]
\caption{Performance Comparison in Gazebo.}
\label{tab:gazebo_comparison}
\centering
\setlength{\tabcolsep}{2.0pt}
\renewcommand{\arraystretch}{1.0}
\resizebox{\columnwidth}{!}{%
\begin{tabular}{@{}lcccccc@{}}
\toprule
Method & SR $\uparrow$ & PT (s) $\downarrow$ & AS (m/s) $\uparrow$ & APT (ms) $\downarrow$ & MSM (m) $\uparrow$ & ASM (m) $\uparrow$ \\
\midrule
MPC-CBF & 60\% & 26.6 & 0.63 & 14.34 & 0.05 & 0.65 \\
MPC-D-CBF & \textbf{100\%} & 22.3 & 0.94 & 16.49 & \textbf{0.44} & 0.88 \\
Proposed & \textbf{100\%} & \textbf{17.2} & \textbf{1.21} & \textbf{0.68} & 0.37 & \textbf{0.95} \\
\bottomrule
\end{tabular}}

\begin{minipage}{0.98\columnwidth}
\footnotesize SR: success rate; PT: passing time; AS: average speed; APT: average planning time; MSM: minimum safety margin; ASM: average safety margin.
\end{minipage}

\end{table}

\begin{figure}[!t]

\centering

\includegraphics[width=\columnwidth]{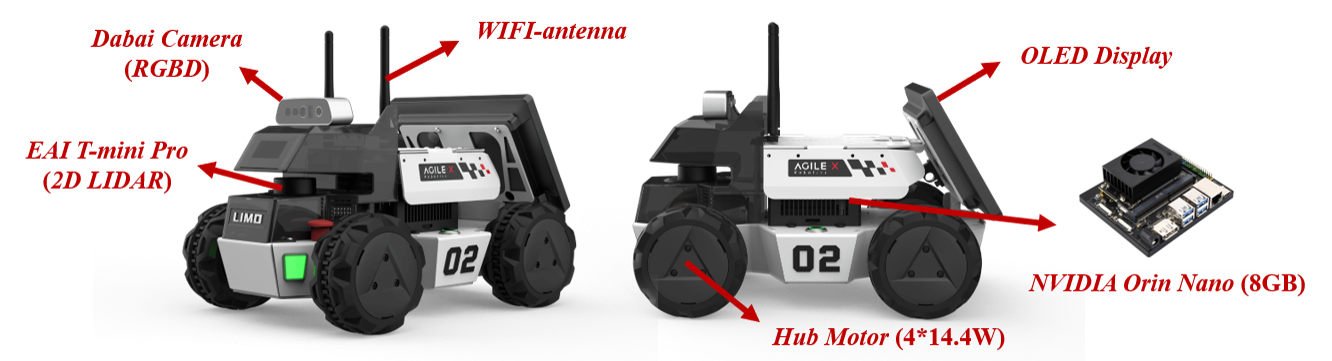}

\setlength{\abovecaptionskip}{0.41\baselineskip}
\caption{Hardware configuration of the AgileX LIMO Pro platform used for real world experiments.}
\label{fig:real_platform}
\vspace{0.1\baselineskip}
\end{figure}

\begin{figure*}[!t]
\centering
\setlength{\abovecaptionskip}{2pt}

\includegraphics[width=0.85\textwidth]{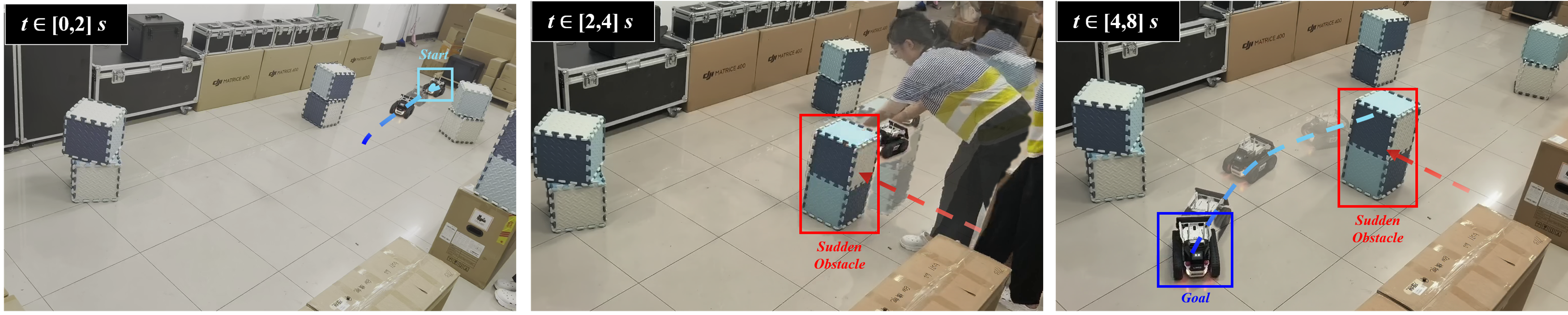}

\caption{Real world sudden obstacle avoidance experiment in an unknown indoor environment.}
\label{fig:real_sudden}

\end{figure*}

\begin{figure}[!t]
\centering

\includegraphics[width=0.65\columnwidth]{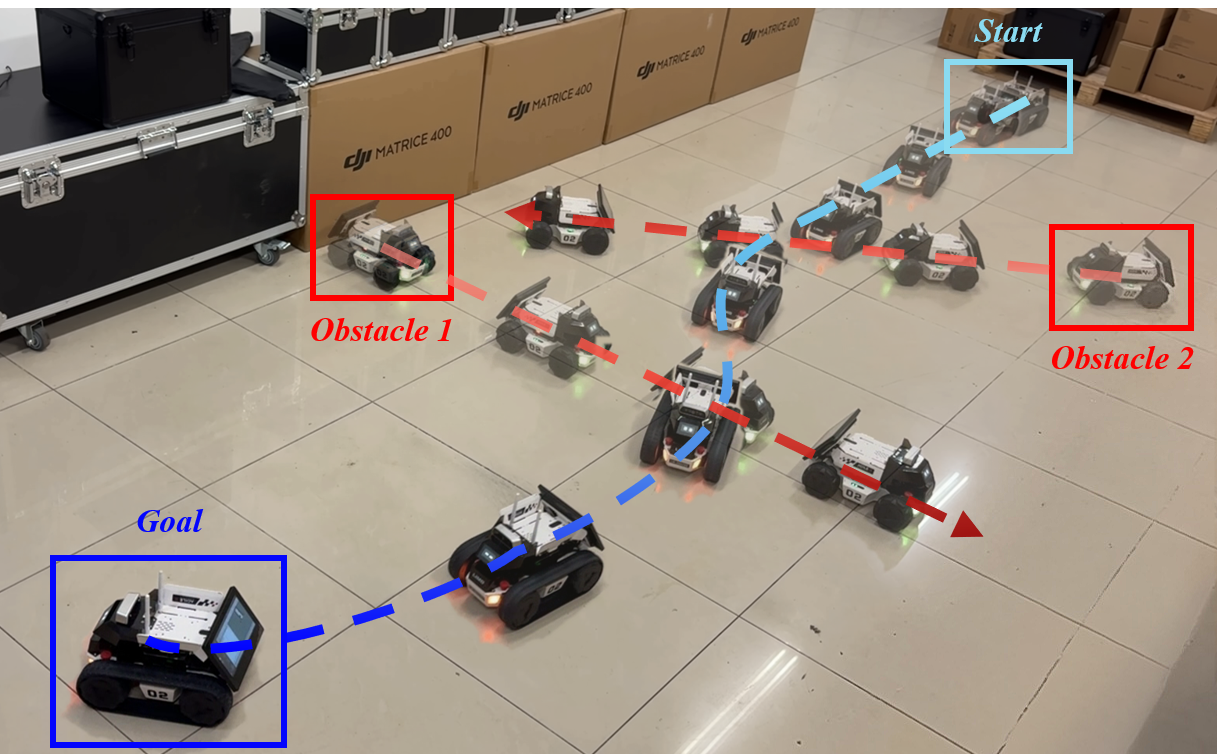}\\[0.3mm]
\makebox[0.75\columnwidth][c]{\footnotesize(a) Two dynamic obstacles crossing laterally}\\[0.5mm]
\includegraphics[width=0.65\columnwidth]{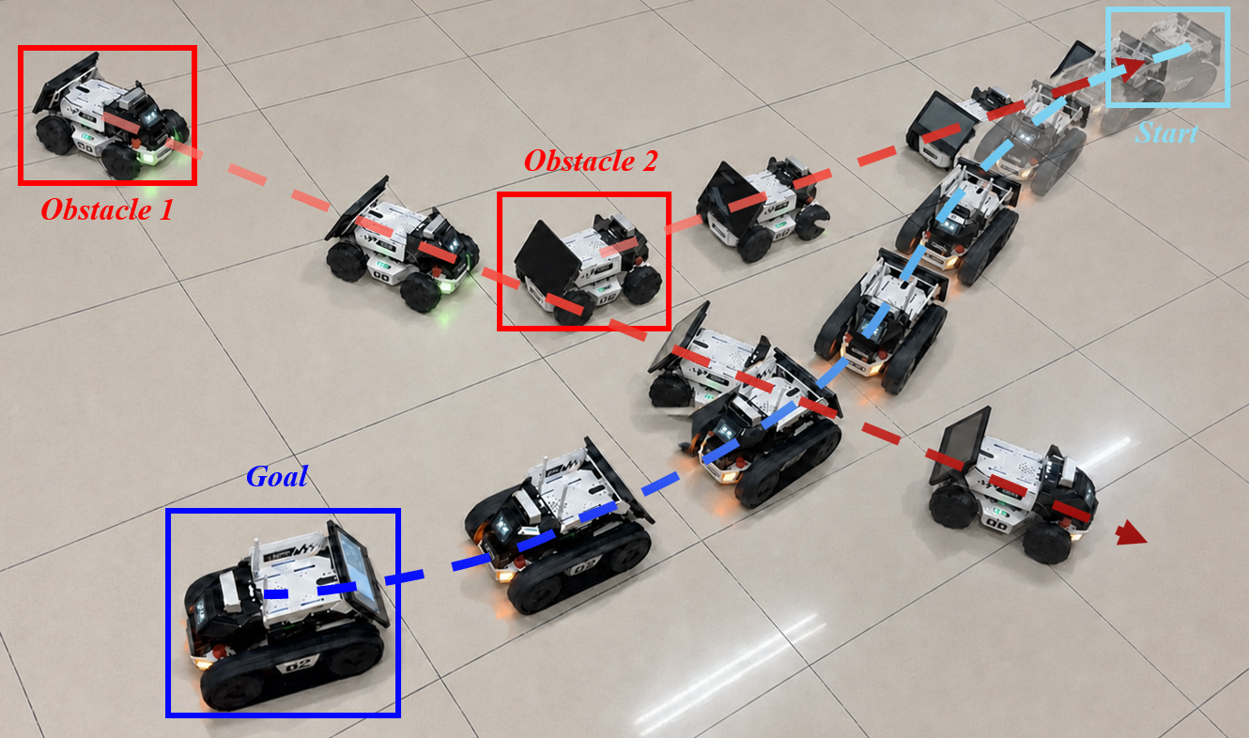}\\[0.3mm]
\makebox[0.75\columnwidth][c]{\footnotesize(b) Two dynamic obstacles with head on and lateral approaches}\vspace{-1.2mm}
\caption{Real world dynamic obstacle avoidance experiment in an unknown indoor environment.}
\label{fig:real_dynamic}

\end{figure}

High-fidelity Gazebo simulations were conducted in a partially enclosed \(20\text{-m}\)-long, \(4.2\text{-m}\)-wide corridor featuring static obstacles and dynamic pedestrians. Implemented in C++ on Ubuntu 20.04 with ROS Noetic, the navigation stack initialized the robot at \((-8~\text{m}, 0~\text{m}, 0~\text{rad})\) with a top speed capped at \(2~\text{m/s}\). Six wooden static obstacles defined central bottlenecks within the \(3.8\text{-m}\)-wide traversable area. Additionally, three pedestrian models (bounding dimensions \(0.3 \times 0.2 \times 1.7~\text{m}\)) traversed reciprocal trajectories at \(1.5~\text{m/s}\).

The proposed framework was benchmarked against two representative safety-critical baselines: MPC-CBF~\cite{zeng2021safety} and MPC-D-CBF~\cite{jian2023dynamic}. While MPC-CBF enforces static CBF constraints within the online MPC optimization, MPC-D-CBF incorporates dynamic CBF constraints to accommodate moving obstacles. Both MPC baselines configured a horizon length of \(N=10\). To ensure a fair comparison, the robot geometry, safety margin, control update frequency, actuation bounds, and CBF gains were kept identical across all schemes, isolating the effects of safety constraint formulation and risk-adaptive mechanisms. Each planner underwent 10 evaluation trials under identical initial states and obstacle trajectories.

As quantified in Table~\ref{tab:gazebo_comparison}, the proposed method and MPC-D-CBF achieved a \(100\%\) completion rate across all 10 trials, whereas MPC-CBF succeeded in only six. The proposed controller cleared the corridor in 17.2~s,outperforming both MPC baselines while securing the highest AS and ASM. Notably, its average planning time (\(0.68~\text{ms}\)) yielded over \(95\%\) computational reduction compared to MPC-D-CBF, underscoring the efficiency of explicit ARB-CBF filtering over repeated non-convex MPC optimization. Although MPC-D-CBF yielded the largest MSM, it suffered from prolonged computational latency and slower overall traversal. As depicted in Fig.~\ref{fig:gazebo_comparison}, the ARB-CBF safety filter dynamically refined nominal control commands during close-range pedestrian interactions without compromising forward progress.

\subsection{Real-World Verification}
\label{sec:real_world}

To validate onboard feasibility, real-world experiments were conducted
on an AgileX LIMO Pro differential drive robot, as shown in
Fig.~\ref{fig:real_platform}. The platform used an NVIDIA Jetson Orin
Nano (8~GB) and an EAI T-mini Pro 2D LiDAR, with localization from wheel
odometry and Cartographer SLAM~\cite{hess2016realtime}. Experiments were
performed in a previously unknown $20~\mathrm{m}\times5~\mathrm{m}$
indoor environment containing five $0.4\times0.4\times0.8$~m static
obstacles, as shown in Fig.~\ref{fig:application}. The ILP profile was
updated for $K=4$ iterations, and the local controller operated at
15~Hz with $v_{\rm init}=0.5$~m/s and $v_{\max}=1.0$~m/s.

\textit{1) Sudden obstacles within the virtual tube:}
As shown in Fig.~\ref{fig:real_sudden}, three static obstacles were
initially arranged around the reference path, and an additional
same-size obstacle was manually moved into the virtual tube during
traversal. The snapshots show online detection, safety correction, and
subsequent passage.

\textit{2) Dynamic obstacles:}
Two identical LIMO Pro robots with an equivalent radius of $0.25$~m
served as dynamic obstacles and moved along straight trajectories at
1.5~m/s. In Fig.~\ref{fig:real_dynamic}(a), both crossed the path
laterally. In Fig.~\ref{fig:real_dynamic}(b), one approached along the
path while the other crossed laterally. The proposed controller safely
handled both interaction patterns.

All scenarios were completed without collision, with an average
planning time of 0.74~ms, average speed of 0.61~m/s, and minimum safety
distance of 0.29~m. These results demonstrate that the proposed framework
maintains lightweight real time execution while providing effective
safety-critical responses to both sudden and moving obstacles in an
unknown environment. Further details are provided in the accompanying
video\footnotemark[1].
\FloatBarrier

\section{Conclusions}

This paper presents a safety-critical iterative learning planning framework tailored for time-efficient navigation of wheeled mobile robots operating in dynamic environments. By coupling iterative learning planning with an anticipatory risk-blended control barrier function (ARB-CBF), the proposed approach guarantees time-optimal trajectory tracking while delivering ultra-lightweight online safety interventions without requiring real-time reoptimization. The framework maintains a low \(O(kN)\) offline computational complexity, facilitating seamless real-time execution on resource-constrained onboard platforms. Extensive simulation benchmarks and real-world experiments on an AgileX LIMO Pro platform confirm the superior traversal efficiency, safety guarantees, and robust performance of the proposed method. Future efforts will focus on learning-based mechanisms for adaptive ILP parameter optimization and extending the framework toward multi-robot cooperative control in high-density dynamic scenes.

\bibliographystyle{IEEEtranTIE}
\bibliography{references}

\end{document}